\documentclass[10pt,twocolumn,letterpaper]{article}

\usepackage{iccv}
\usepackage{times}
\usepackage{epsfig}
\usepackage{graphicx}
\usepackage{amsmath}
\usepackage{amssymb}

\usepackage[breaklinks=true,bookmarks=false]{hyperref}  

\iccvfinalcopy 

\def\iccvPaperID{****} 

\ificcvfinal  \fi

\usepackage{booktabs}

\newcommand{\dc}{D$^2$-City}
\newcommand{\bdc}{D$\mathbf{^2}$-City}

\newcommand{\NTotal}{$11,211$}

\begin{document}

\title{\bdc: A Large-Scale Dashcam Video Dataset of Diverse Traffic Scenarios} 

\author{
Zhengping Che$^{1}$, Guangyu Li$^{1}$, Tracy Li$^{2}$, \\
Bo Jiang$^{1}$, Xuefeng Shi$^{1}$, Xinsheng Zhang$^{1}$, Ying Lu$^{1}$,
Guobin Wu$^{2}$, Yan Liu$^{3}$, Jieping Ye$^{1}$\\
$^{1}$ DiDi AI Labs, Didi Chuxing, Beijing, China\\
$^{2}$ Didi Chuxing, Beijing, China\\
$^{3}$ University of Southern California, Los Angeles, CA, USA\\
{\tt\small \{chezhengping, maxliguangyu, liquntracy\}@didiglobal.com}\\
{\tt\small \{scottjiangbo, shixuefeng, zhangxinsheng, yinglu, wuguobin, yejieping\}@didiglobal.com}\\
{\tt\small yanliu.cs@usc.edu}
}

\maketitle

\begin{abstract}

Driving datasets accelerate the development of intelligent driving and related computer vision technologies, while substantial and detailed annotations serve as fuels and powers to boost the efficacy of such datasets to improve learning-based models.
We propose {\dc}, a large-scale comprehensive collection of dashcam videos collected by vehicles on DiDi's platform.
{\dc} contains more than $10,000$ video clips which deeply reflect the diversity and complexity of real-world traffic scenarios in China.
We also provide bounding boxes and tracking annotations of 12 classes of objects in all frames of $1000$ videos and detection annotations on keyframes for the remainder of the videos.
Compared with existing datasets, {\dc} features data in varying weather, road, and traffic conditions and a huge amount of elaborate detection and tracking annotations.
By bringing a diverse set of challenging cases to the community, we expect the {\dc} dataset will advance the perception and related areas of intelligent driving. 
\end{abstract}

\section{Introduction}

The recent boom of publicly available driving video datasets has enhanced many real-world computer vision applications such as scene understanding, intelligent transportation, surveillance, and intelligent driving.
Given the huge amount of raw video data collected, accurate object detection and multi-object tracking is one of the key elements and vital challenges in these applications, as temporal and dynamic information is rich in video sequences and may also bring extra difficulties.
Most existing driving datasets have not paid enough attention to these tasks or were not able to fully address them.
The KITTI Vision Benchmark Suite~\cite{geiger2013vision} proposed several benchmark tasks on stereo, optical flow, visual odometry, 3D object detection and 3D tracking.
The CityScapes Dataset~\cite{cordts2016cityscapes} aimed at pixel-level and instance-level semantic labeling tasks and provided $25,000$ fully or weakly annotated images.
The numbers of annotated data of these datasets are relatively small, which makes it difficult for models to learn and understand complex scenarios in the real world.
ApolloScape~\cite{huang2018apolloscape} provided more than $140,000$ image frames with pixel-level annotations, and frames for segmentation benchmark were extracted from a limited set of videos.
BDD100K~\cite{yu2018bdd100k} collected over $100,000$ video clips and provide several types of annotations for key frames of each video. Though raw video sequences were provided, the annotations were only for one single-frame from each video and all data were collected in US where the traffic conditions are different from other places such as east Asia.
In short, large-scale datasets that fully reflect the complexity of real-world scenarios and present rich video-level annotations, such as tracking of road objects, are still in great demand.

We propose the {\dc} dataset, which is a large-scale driving video dataset collected in China.
{\dc} provides more than $10,000$ videos recorded in 720p HD or 1080p FHD from front-facing dashcams, with detailed annotations for object detection and tracking.
Compared with existing datasets, {\dc} demonstrates greater diversity, as data is collected from several cities throughout China and features varying weather, road, and traffic conditions.
{\dc} pays special attention to challenging and actual case data, which is recorded by dashcams equipped in passenger vehicles on DiDi's platform. By bringing a diverse set of challenging real cases with detailed annotations to the community, we expect {\dc} will encourage and inspire new progress in the research and applications of perception, scene understanding, and intelligent driving. {\dc} is one of the first, if not the only, public large-scale driving datasets available for both detection and tracking tasks in real circumstance. In summary:
\begin{itemize}
  \item We release {\NTotal} driving videos, which are about one hundred hours long in total. The data collection fully respects the diversity and complexity of real traffic scenarios in China.
  \item For $1000$ of the collected videos with more than $700,000$ frames, we annotate detailed detection and tracking information of on-ground road objects, including the bounding box coordinates, tracking IDs, and class IDs for $12$ classes.
      We also provide detection annotations in keyframes for the remainder of the videos in stages.
  \item Based on our videos and annotations, we provide training, validation, and testing subsets and three tasks including object detection, multi-object tracking, and large-scale detection interpolation.
\end{itemize}

The {\dc} dataset is available online at \url{https://outreach.didichuxing.com/d2city}.

\section{Dataset}
\begin{figure}[!htbp]
    \begin{center}
        \includegraphics[width=0.815\linewidth]{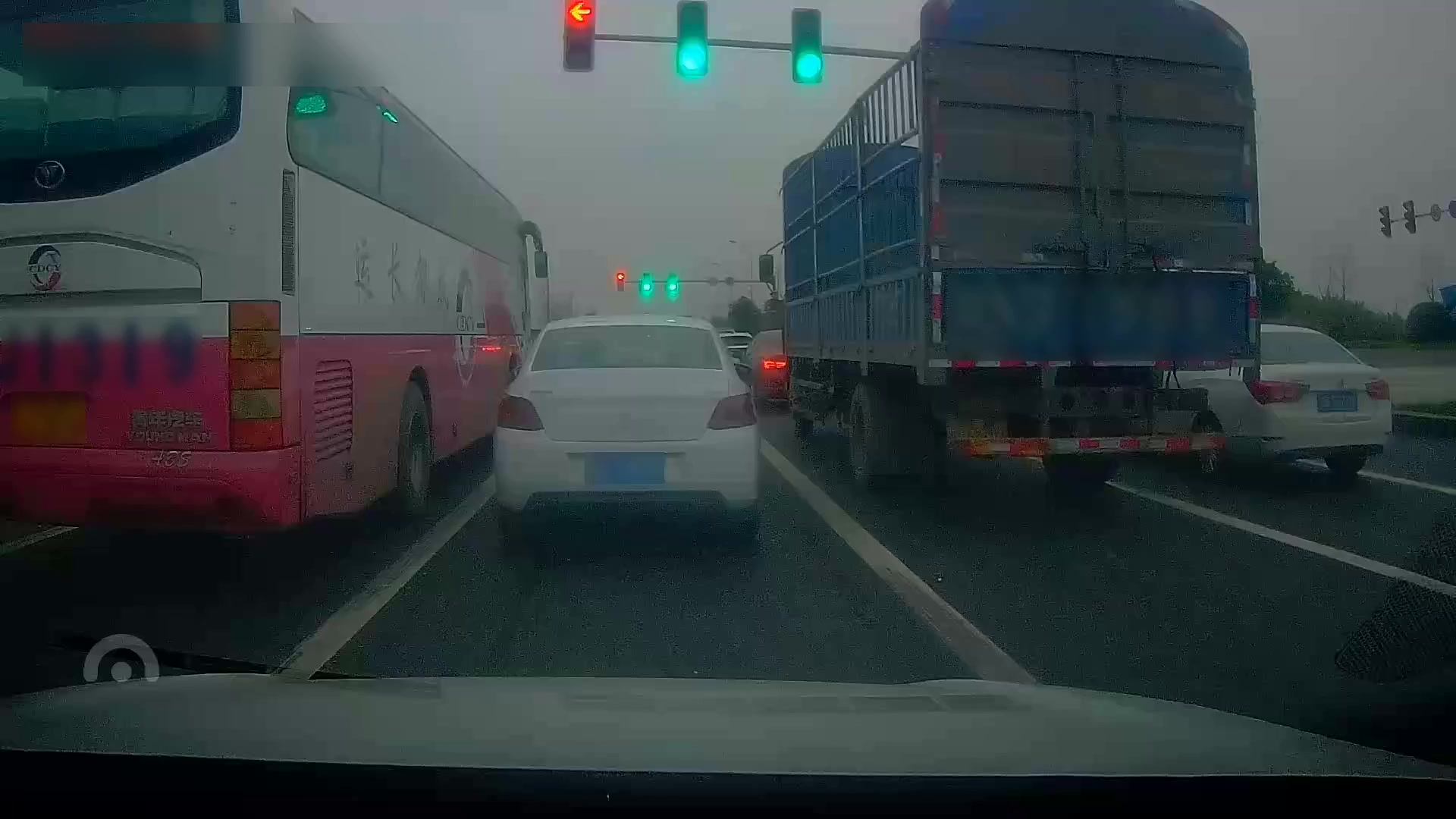} \\ \vspace{0.05in}
        \includegraphics[width=0.815\linewidth]{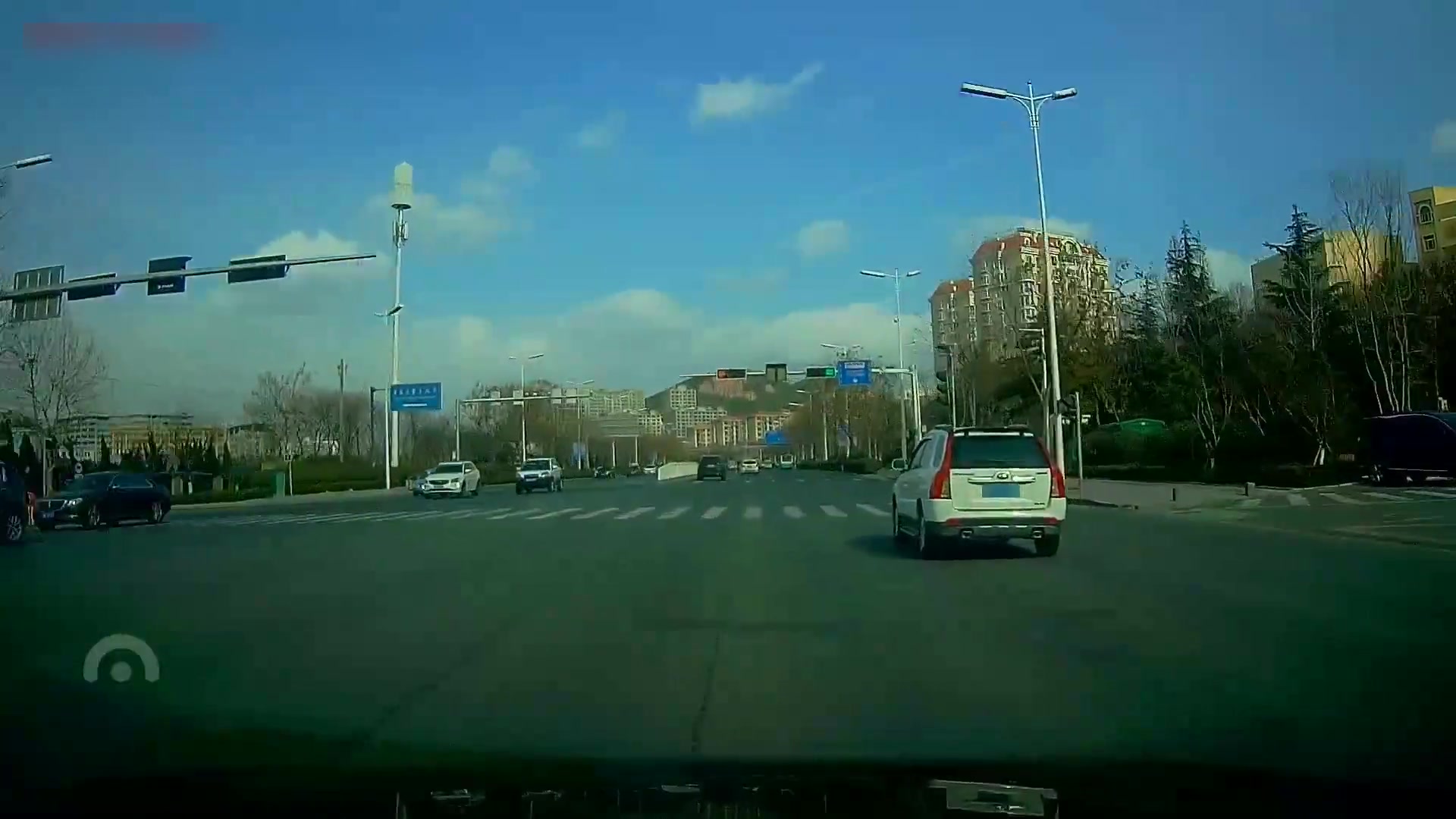} \\ \vspace{0.05in}
        \includegraphics[width=0.815\linewidth]{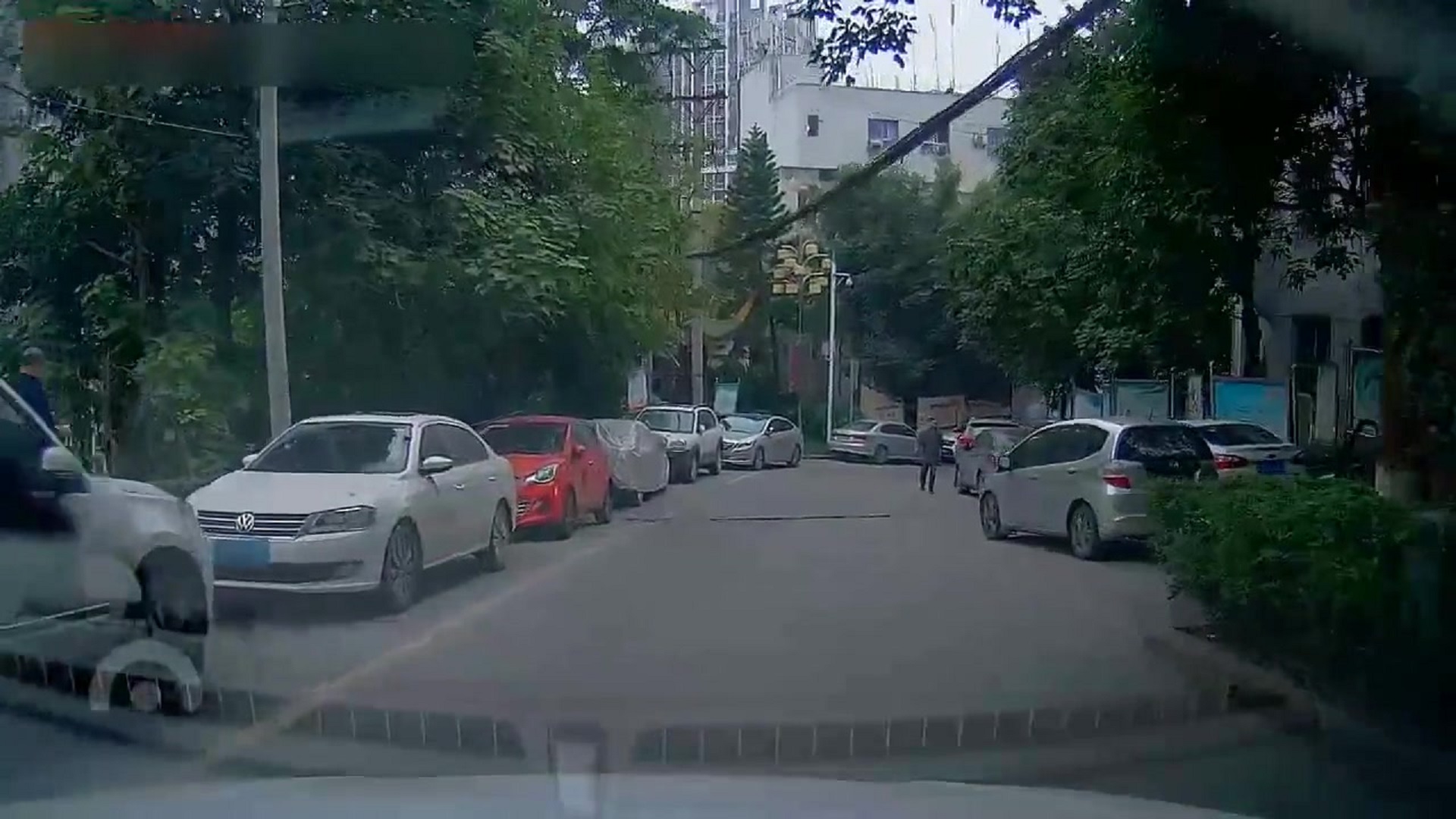} \\ \vspace{0.05in}
        \includegraphics[width=0.815\linewidth]{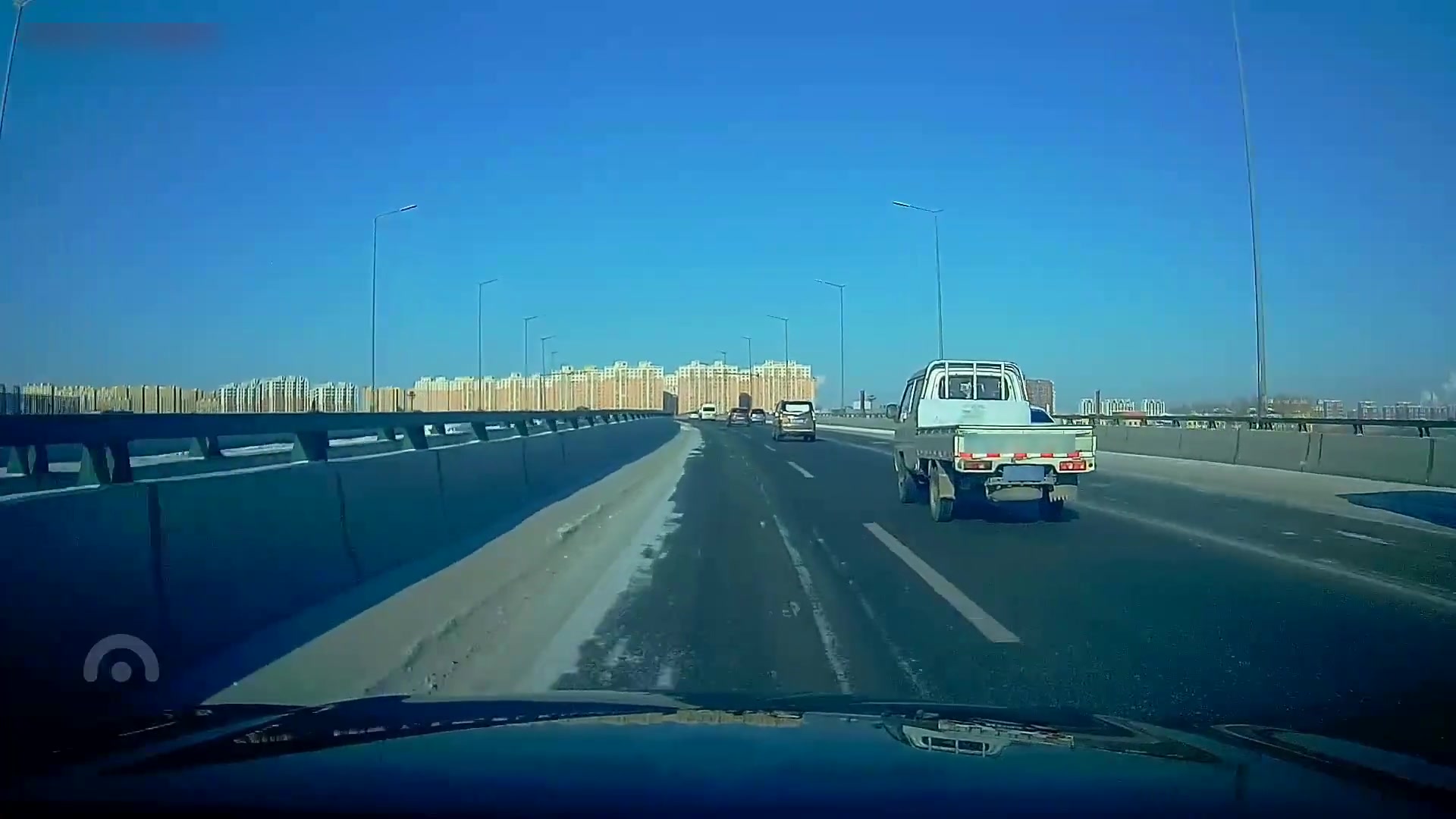} \\ \vspace{0.05in}
        \includegraphics[width=0.815\linewidth]{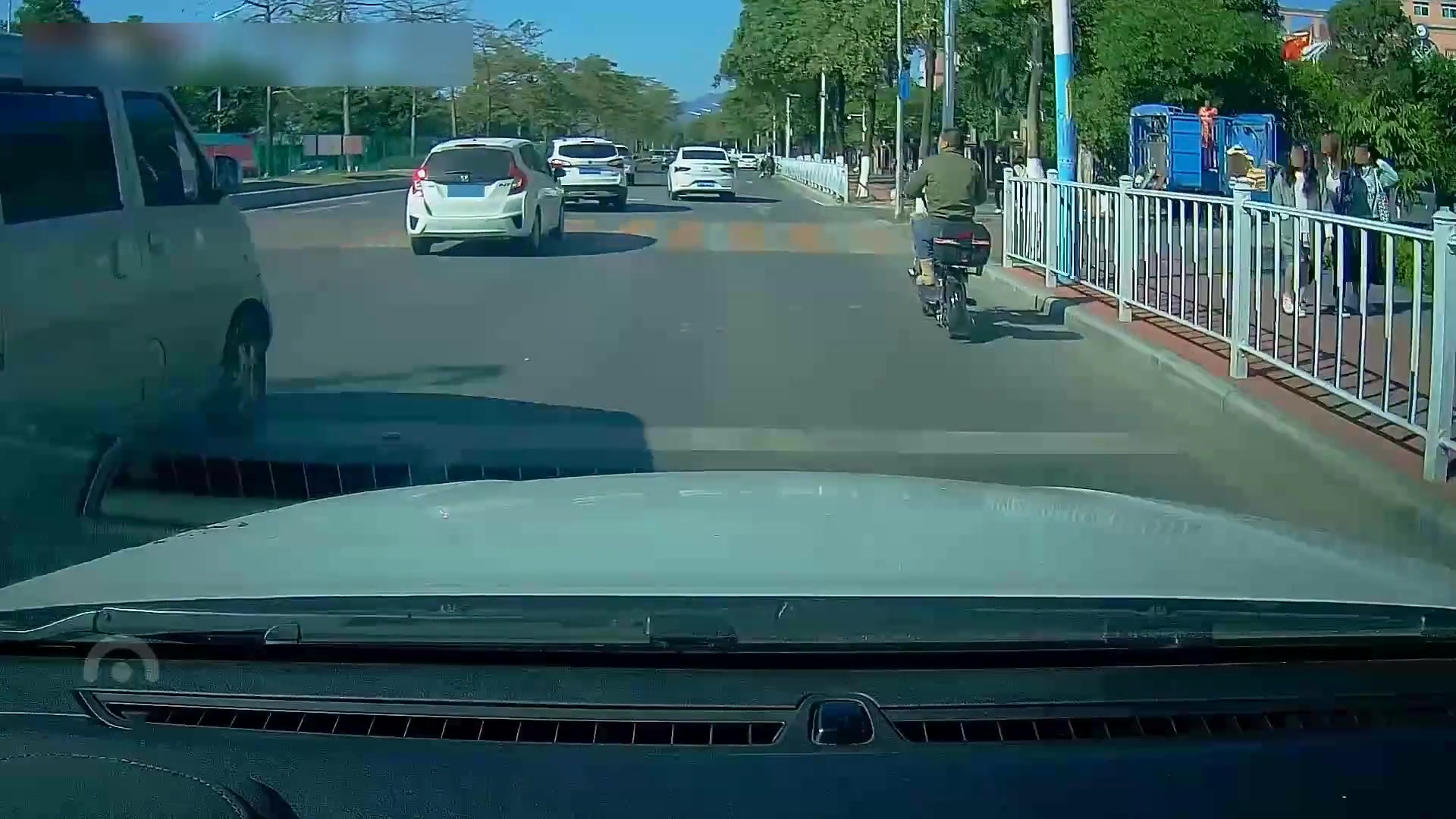} \\
    \end{center}
    \caption{A list of sample video frames from the {\dc} data collection. {\dc} covers diverse real-world traffic scenarios in China, such as traffic congestions, crowded crossroads, narrow alleys, road constructions, and scenes with large volumes of non-motor vehicles and pedestrians.}
    \vspace{-0.2in}
    \label{fig:long}
\end{figure}

\subsection{Data Collection}

Our data were all collected in a crowd-source way from vehicles on DiDi's platform.
Hundreds of thousands of vehicles are operated and online everyday all over China with dashcams which can capture front-facing driving videos, and the number keeps increasing steadily.
Therefore, the coverage and the diversity of our data collection is well guaranteed.
In {\dc}, we focused on five cities and selected a few vehicles in these cities based on their online time, main driving area, and the quality of their sampled image frames.
Due to bandwidth and memory constraint and data traffic cost, we only recorded and uploaded no more than 2 minutes of video in one hour.
All videos were recorded in 30 second-long clips at 25Hz.
As vehicles are equipped with different model types of cameras and the network conditions usually change, the collected videos have different resolutions of 720p or 1080p and different bitrates between 2000kbps and 4000kbps in its raw form. We kept its original resolutions and bitrates to maintain the diversity of our data.

\subsection{Data Selection}
We collected more than one thousand hours of videos.
Some of the videos seemed not to be proper or useful, due to data quality, diversity, or other issues, for research and application purposes.
Therefore, we applied detailed and tailored data selection in terms of video quality, scenario diversity, and information security.

\paragraph{Video quality}
As videos were recorded by low-cost, daily-use dashcams which are not specifically designed for collecting high-quality data but just for routine recording, they varied a lot in quality.
After examining some sample frames and pre-collected videos, we found several common issues, including image blur, light-related problems, reflection on the front windshield, and wrong angle of dashcam shot.
Extreme weather conditions such as rain, snow, or haze decreased video qualities temporarily.
Sometimes blurred image frames also came from uncleaned camera shots or front windshields.
Light-related problems, including dim light, overexpose, and strong glare, usually took place at particular time or in particular areas.
Some drivers often put decorations, tissue boxes, portable mobile devices, or other things in the front of their vehicles, and the reflections of these things on the windshield became interference.
A few dashcams had relatively low angle, tilted angle or partially occluded outside view at times.
Though videos collected by these dashcams were still useful for other general purposes, they might be less valuable for the research of intelligent driving, at least at the current stage.
In practice, we took a 3-step action to keep videos with satisfying quality.
Before we collected videos, we sampled frames from each device and selected devices which constantly produced relatively high quality videos.
Second, we carried out spot check on sampled frames everyday to ensure the quality of videos collected from each device on that day.
Finally, we took manpower to skim all remaining videos quickly as a sanity check.

\paragraph{Scenario diversity}
We expect {\dc} not only contains large amount of data but also fully demonstrates the diversity in data.
Therefore, we followed a few simple rules to avoid capturing tedious or less valuable video clips.
We excluded videos where the ego-vehicle stayed for most of the 30-second period.
We also skipped the next video clip if the previous one was selected, and therefore no consecutive videos are selected.
Finally, we avoided to collect too many videos from the same device even if the videos satisfied other criteria, in order to enlarge the diversity of driving behaviours and routes in {\dc} by including more vehicles and drivers in our dataset.

\paragraph{Information security}
Drivers might put some personal items under the front windshield. Sometimes the dashcam may fully or partially capture those items directly or from the reflections on the front windshield.
We excluded such videos to ensure the quality of the videos and protect privacy.
We also carefully excluded places where taking photos or videos is prohibited or discouraged from our video collections if any.
YOLOv3~\cite{redmon2018yolov3} and DF$^2$S$^2$~\cite{tian2018learning} models were used to detect and blur license plates and pedestrians' faces for privacy protection.
Additionally, we blurred all timestamps embedded in the top-left corners of raw video frames.
These steps hid few objects such as traffic lights and barely impacted instances of the object classes that we annotated.

\subsection{Data Specification}
\begin{figure}[htb]
    \begin{center}
        \includegraphics[width=0.58\columnwidth]{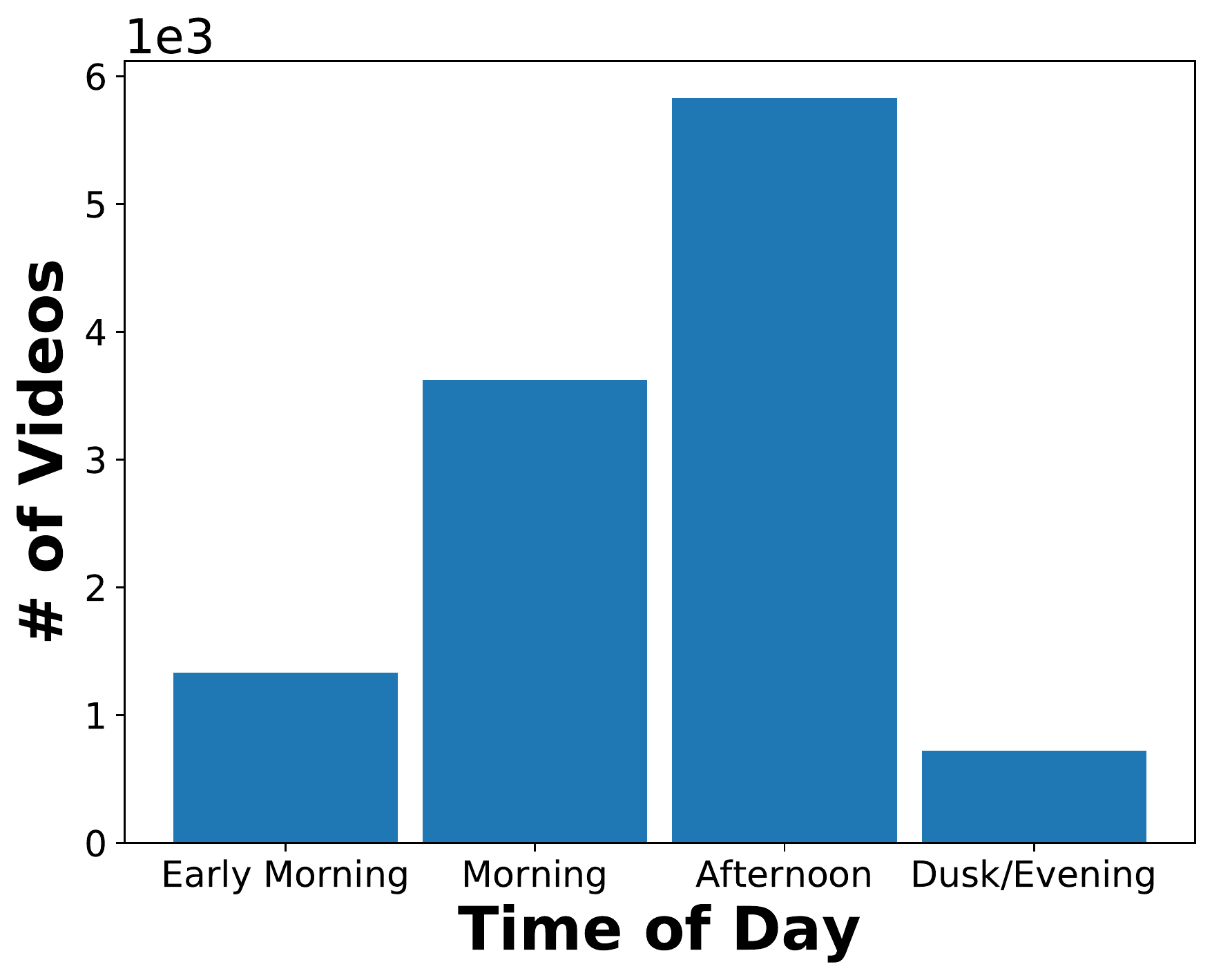}
    \end{center}
    \caption{Distribution of videos in different time of day.}
    \label{fig:time-of-day}
\end{figure}

Following the criteria mentioned above, we selected {\NTotal} videos collected by about $500$ vehicles operated on DiDi's platform in 5 cities in China.
$8037$ videos were recorded in 1080p FHD and the rest $3174$ videos were in 720p HD.
Figure~\ref{fig:time-of-day} shows the distribution of all selected videos in different time of day. We mainly collected videos in the daytime as videos are in better quality especially in terms of light, while about one fifth of all videos are in the early morning or late evening.

\begin{table}[htb]
    \begin{center}
        \begin{tabular}{c|c|c}
            \toprule
            Highway & Arterial Road & Sub-Arterial Road \\ \cmidrule{1-3}
            $938$ & $401$ & $2610$ \\
            \midrule
            Ramp & Side Road & Toll Road \\ \cmidrule{1-3}
            $408$ & $700$ & $317$ \\
            \bottomrule
        \end{tabular}
    \end{center}
    \caption{Numbers of videos with different road types and classes.}
    \label{tab:road-type}
\end{table}

\begin{figure}[htb]
    \begin{center}
        \includegraphics[width=0.58\columnwidth]{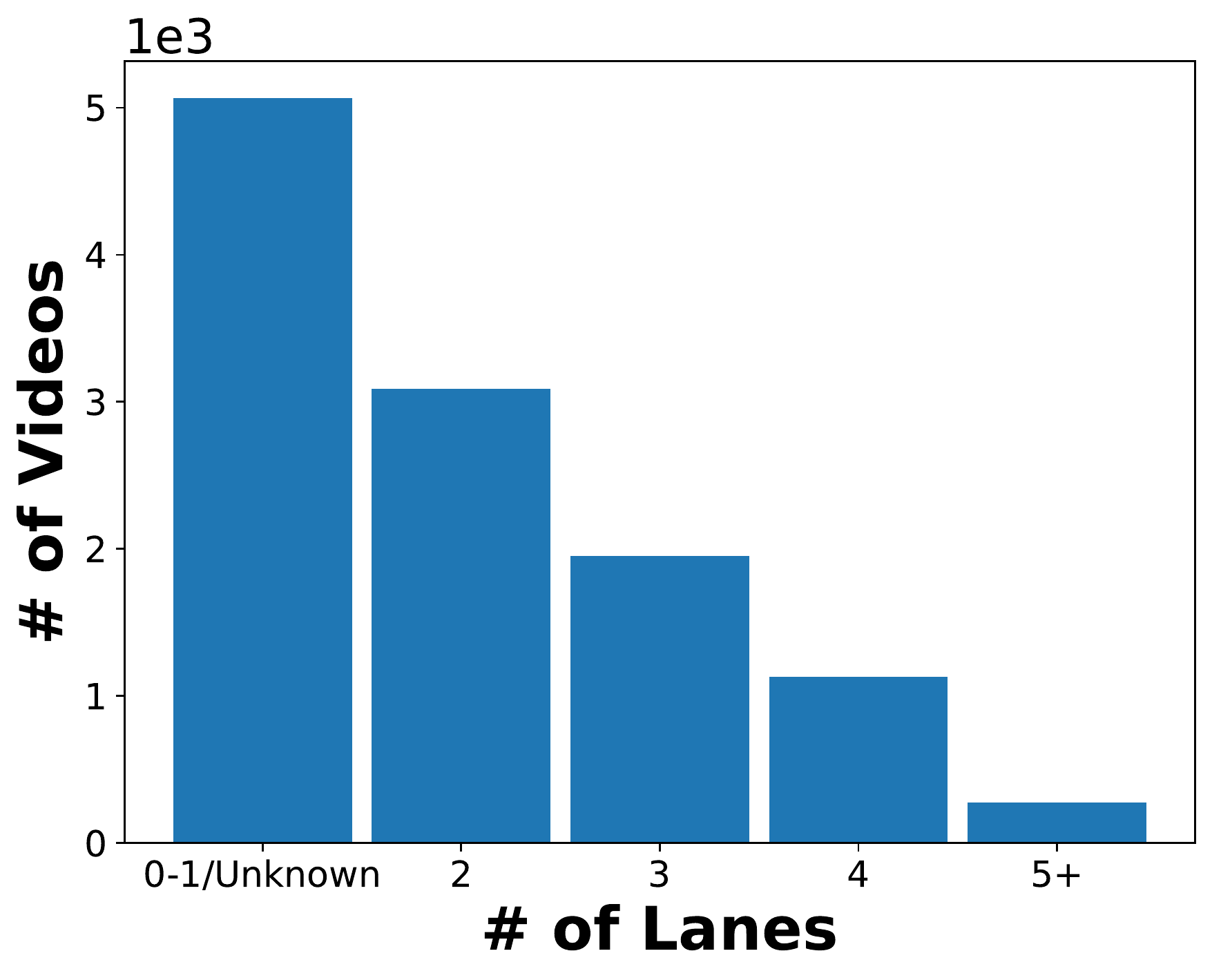}
    \end{center}
    \caption{Statistics of the number of lanes in the driving direction in all videos.}
    \label{fig:lanes}
\end{figure}

Among all selected videos, $6159$ videos were collected in urban areas, and others were collected in suburbs or other areas.
As shown in Table~\ref{tab:road-type}, {\dc} covers a variety of road types and classes.
The statistics of average numbers of lanes in the driving direction is shown in Figure~\ref{fig:lanes}. In more than half of the scenarios we collected, there are at least 2 lanes in the driving direction.

\begin{figure}[htb]
    \begin{center}
        \includegraphics[width=0.58\columnwidth]{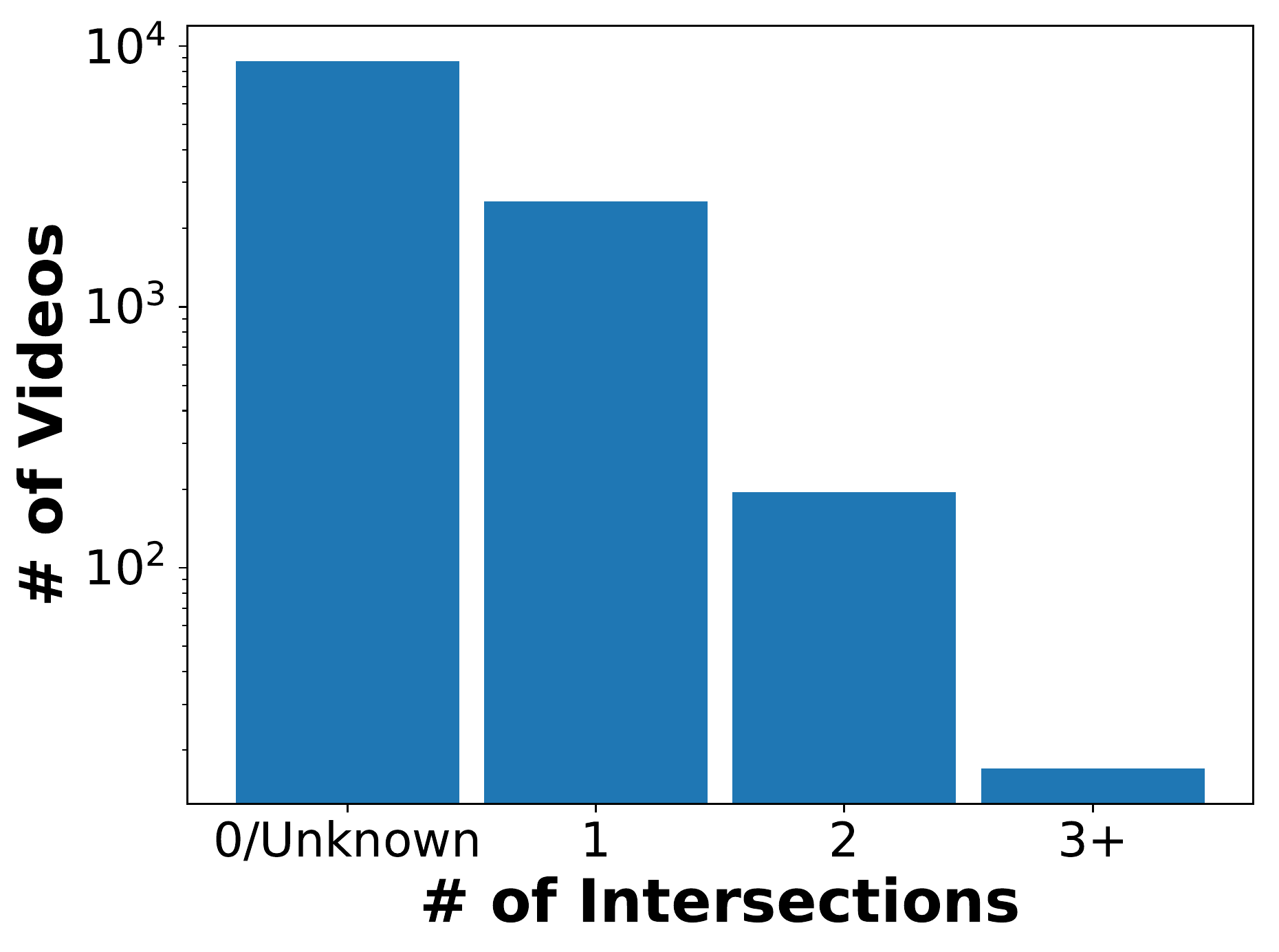}
    \end{center}
    \caption{Statistics of the numbers of intersections the ego-vehicles passed in all videos.}
    \label{fig:dist-intersect}
\end{figure}

\begin{table}[htb]
    \begin{center}
        \begin{tabular}{c|c|c|c}
            \toprule
            Straight & Right Turn & Left Turn & U-Turn\\ \cmidrule{1-4}
            $2123$ & $301$ & $281$ & $16$ \\
            \bottomrule
        \end{tabular}
    \end{center}
    \caption{Number of videos of different driving directions at intersections.}
    \label{tab:intersect}
\end{table}

\begin{figure}[b!]
    \onecolumn
    \begin{center}
        \hfill
        \includegraphics[width=0.305\linewidth]{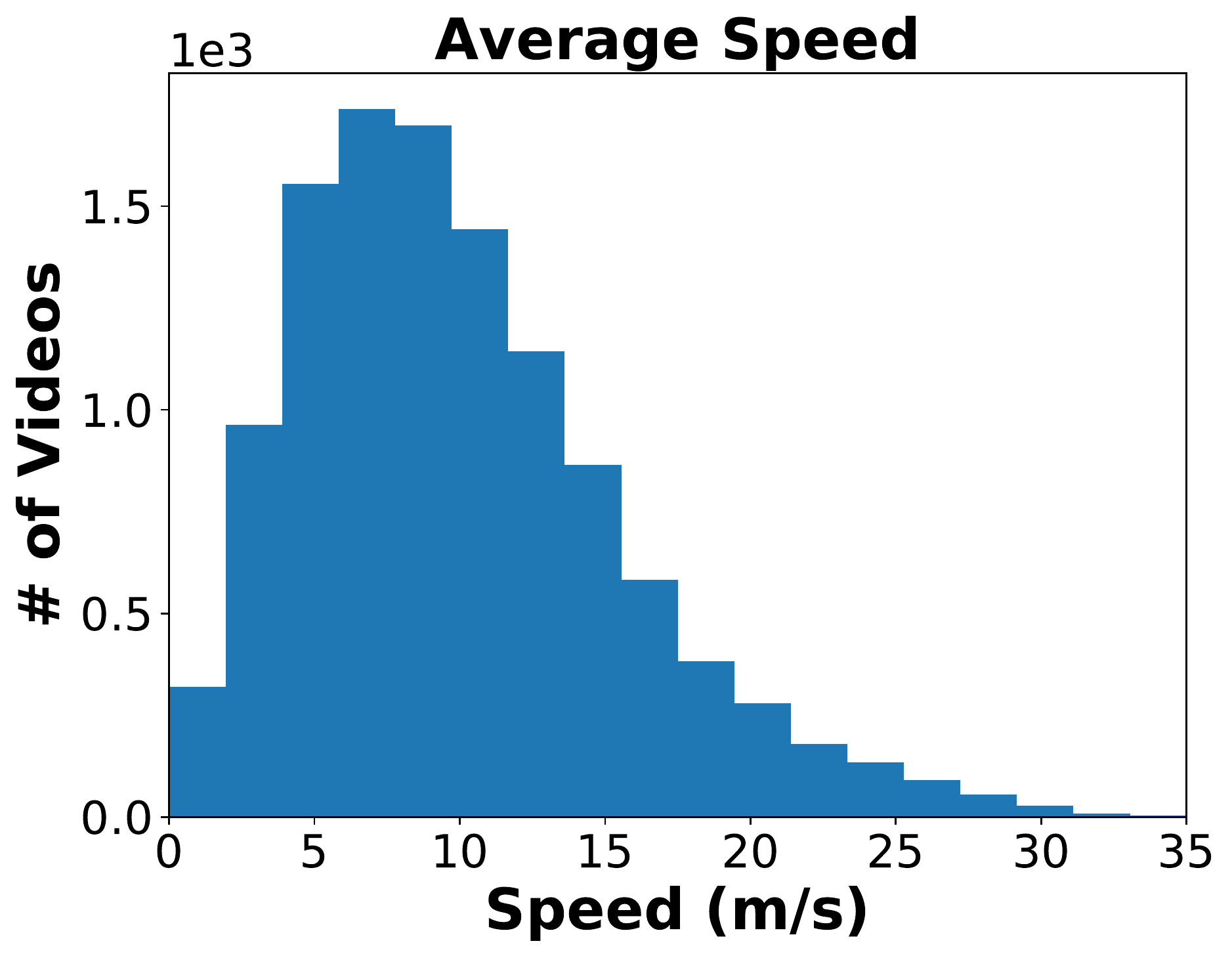} \hfill
        \includegraphics[width=0.305\linewidth]{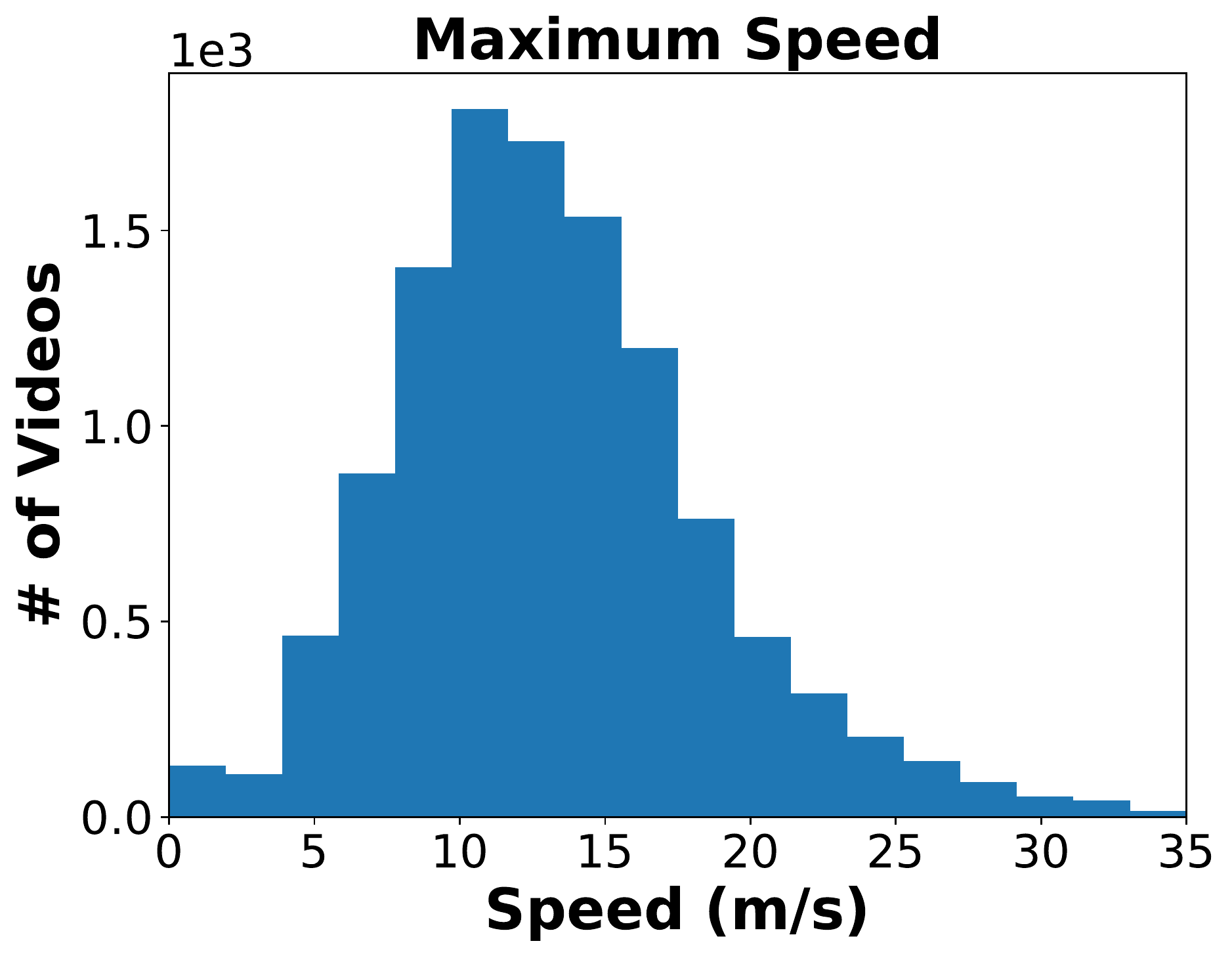} \hfill
        \includegraphics[width=0.305\linewidth]{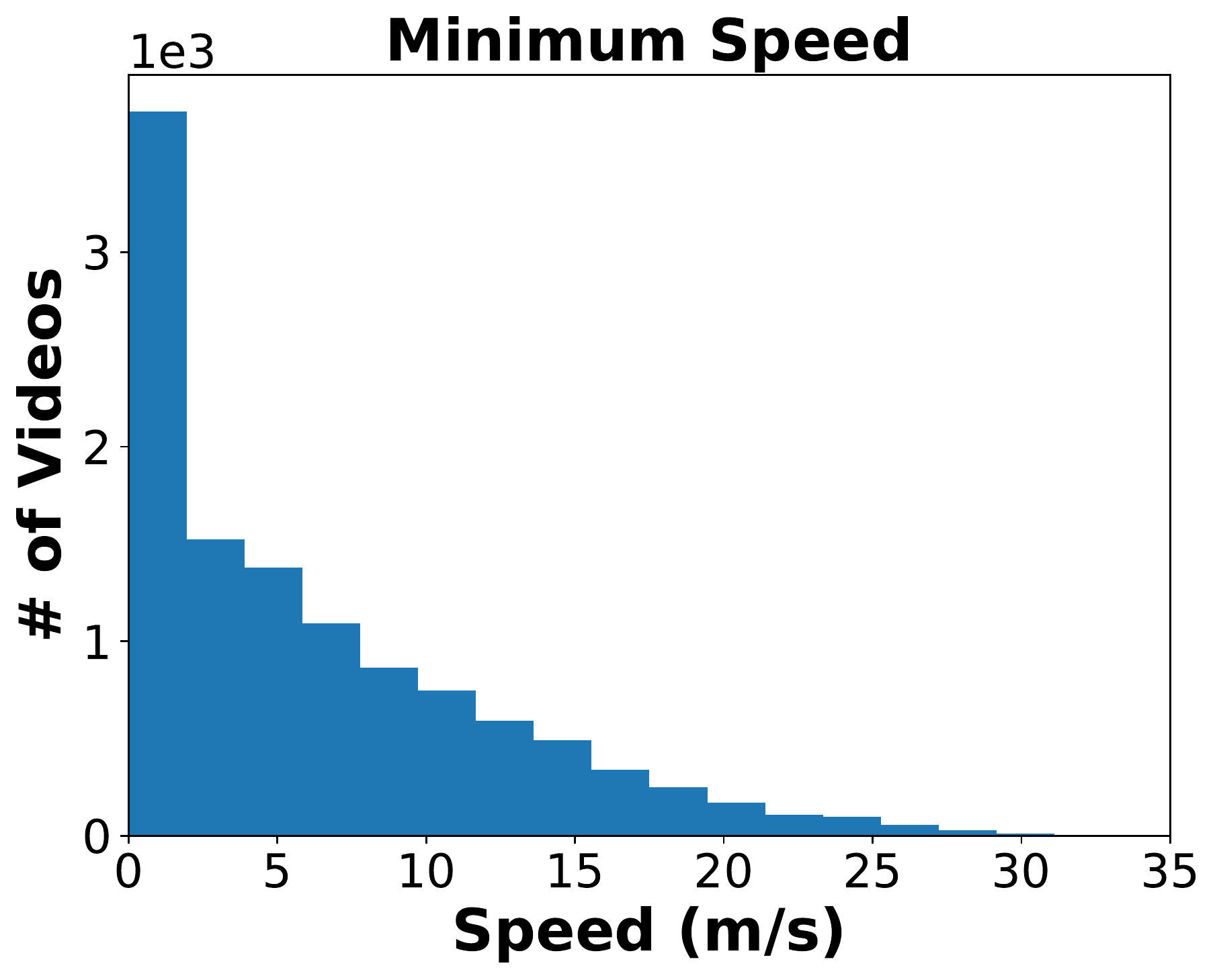} \hfill $\ $
    \end{center}
    \caption{Distributions of the average, maximum, and minimum speed of the ego-vehicles in all videos.}
    \label{fig:dist-speed}
    \twocolumn
\end{figure}

On average, ego-vehicles passed $0.26$ intersections in one 30-second video clip as shown in Figure~\ref{fig:dist-intersect}.
In other words, the ego-vehicles passed about 1 intersection every other minute.
Table~\ref{tab:intersect} shows the distributions of driving behaviors at intersections in all videos.
In about $79.2\%$ cases the driver went straight through the intersection.
The occurrences of right turns at intersections are slightly more than those of left turns.
The average speed of the ego-vehicles in these videos is $9.84$m/s ($35.4$km/h), and the average maximum and minimum speeds of the ego-vehicles during video clips are $13.70$m/s ($49.3$km/h) and $5.94$m/s ($21.4$km/h), respectively. The distributions of the average, maximum, and minimum speeds of all video clips are shown in Figure~\ref{fig:dist-speed}.

\section{Annotation}
\subsection{Annotation and Task Design}

Detailed annotations are costly and time-consuming, especially when we need both per-frame bounding boxes, attributes, and inter-frame tracking labels.
After investigating and learning from several existing annotation tools, we developed an efficient customized annotation platform based on Computer Vision Annotation Tool (CVAT)~\cite{springerlink:10.1007/s11263-012-0564-1}.
To balance labeling quality and efficiency, all annotations were created manually or by frame propagation and mean-shift interpolation within very short time ranges with manual adjustments.
We also found that loading predictions from well-trained object detection and tracking models as pre-annotations may further speedup the annotation procedure as mentioned in existing annotation systems~\cite{yu2018bdd100k}. However, in order to guarantee the quality and purity of the ground-truths, we did not use any learning-based models to get pre-annotations in labeling {\dc}.
Our annotation platform also supports labeling quality inspection and labor monitoring for each task and each annotator, so that we could track labeling quality and progress and conduct necessary adjustments and rework in a timely manner.

In {\dc}, we provided annotations of 12 classes of road objects, named as \textit{car}, \textit{van}, \textit{bus}, \textit{truck}, \textit{person}, \textit{bicycle}, \textit{motorcycle}, \textit{open-tricycle}, \textit{closed-tricycle}, \textit{forklift}, \textit{large-block}, and \textit{small-block}.
Similar to ApolloScape~\cite{huang2018apolloscape}, we paid special attention to three-wheel vehicles which are common on roads in China.
In addition, we further provided two classes for open tricycles (\textit{open-tricycle}) and closed-door tricycles (\textit{closed-tricycle}) because they have quite different appearances.
Same as other vision datasets~\cite{everingham2010pascal,yu2018bdd100k}, we also added the truncation and occlusion flags for each bounding box in three levels (\textit{no}, \textit{partially}, and \textit{almost}).
Class examples and some specific labeling rules are discussed in Appendix~\ref{sec:class-def}.

\begin{figure}[!b]
\vspace{2.2in}
\end{figure}

We annotated bounding box and tracking annotations of 12 classes of road objects on all frames in $1000$ videos in {\dc}. we split those videos into training ($700$ videos), validation ($100$ videos), and test ($200$ videos) sets.
The annotations on the training and validation sets are released publicly. This part of {\dc} is supposed to be used as a large-scale benchmark dataset for road object detection and tracking tasks.
For the remainder of videos in {\dc}, we released bounding box annotations of 12 classes on some key frames and designed a large-scale detection interpolation task on those videos.
We hope plentiful raw video data with key annotations in {\dc} can inspire novel and efficient solutions to obtain accurate detection results by properly leveraging algorithms and manpower in practice.

\subsection{Label Statistics}
We show the statistics of detection and tracking annotations on the training set.
The numbers on the validation and test sets follow similar patterns.

\begin{figure}[b!]
    \begin{center}
        \hfill
        \includegraphics[width=0.47\linewidth]{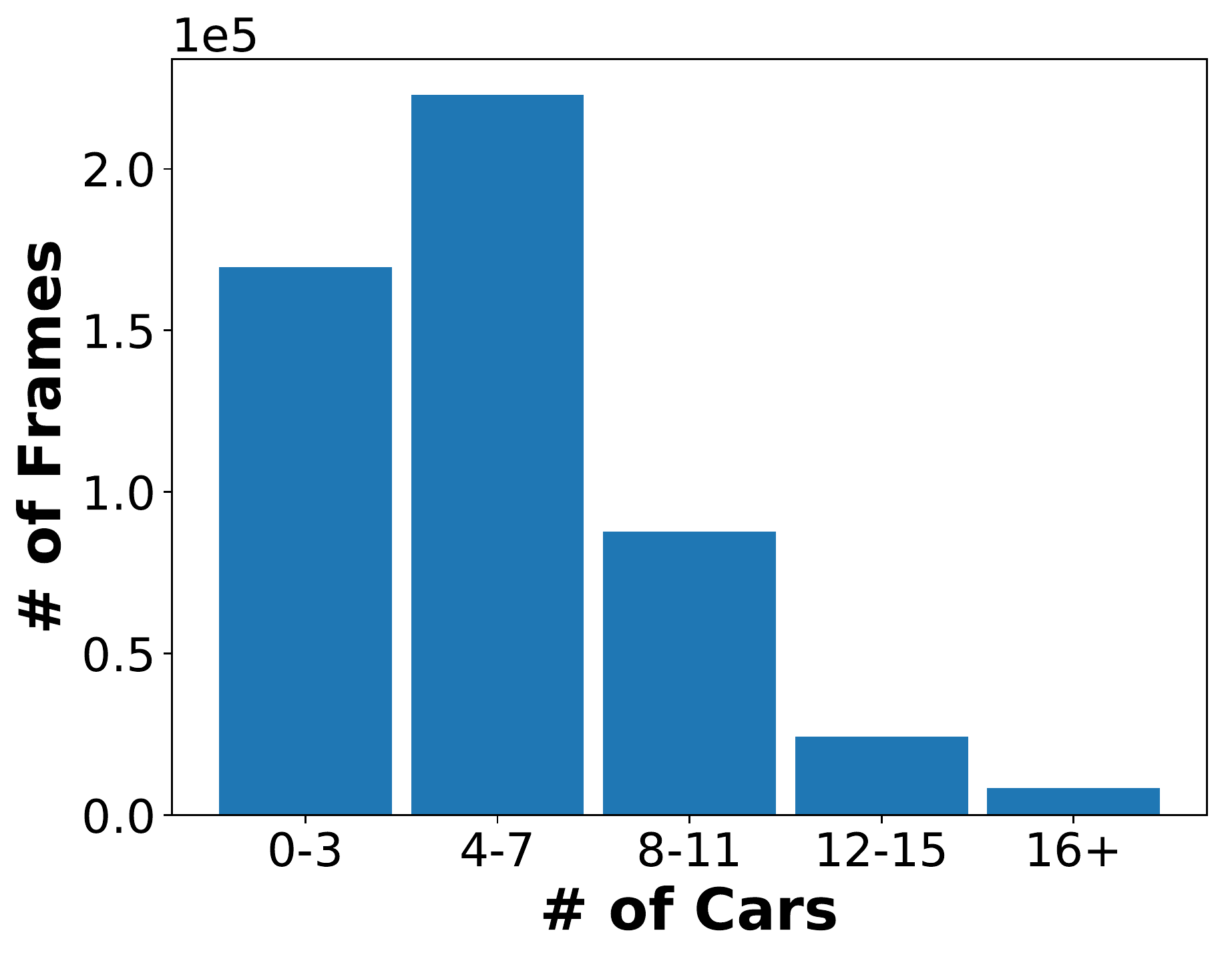}  \hfill $\ $ \\ \vspace{0.1in}
        \hfill
        \includegraphics[width=0.47\linewidth]{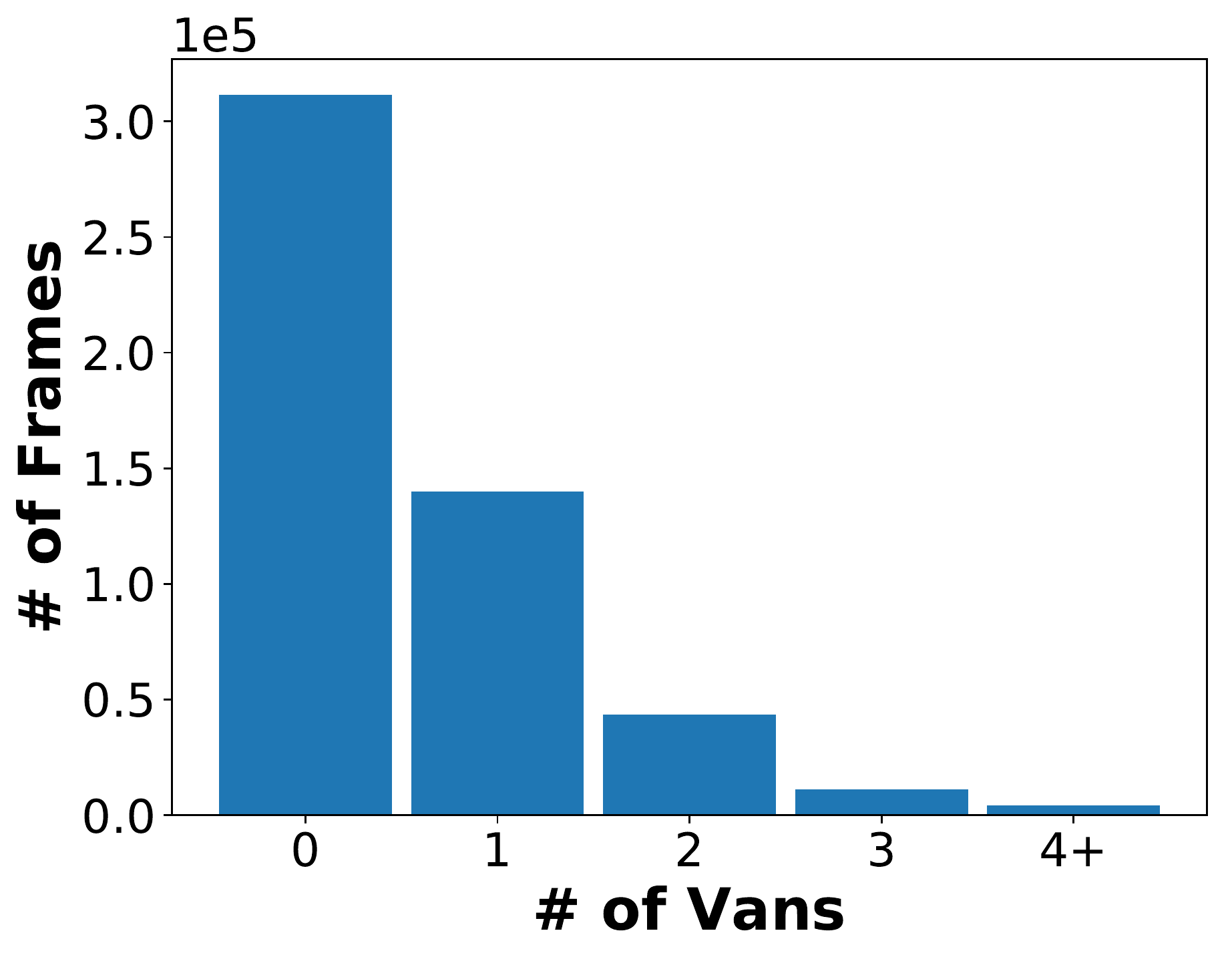} \hfill
        \includegraphics[width=0.47\linewidth]{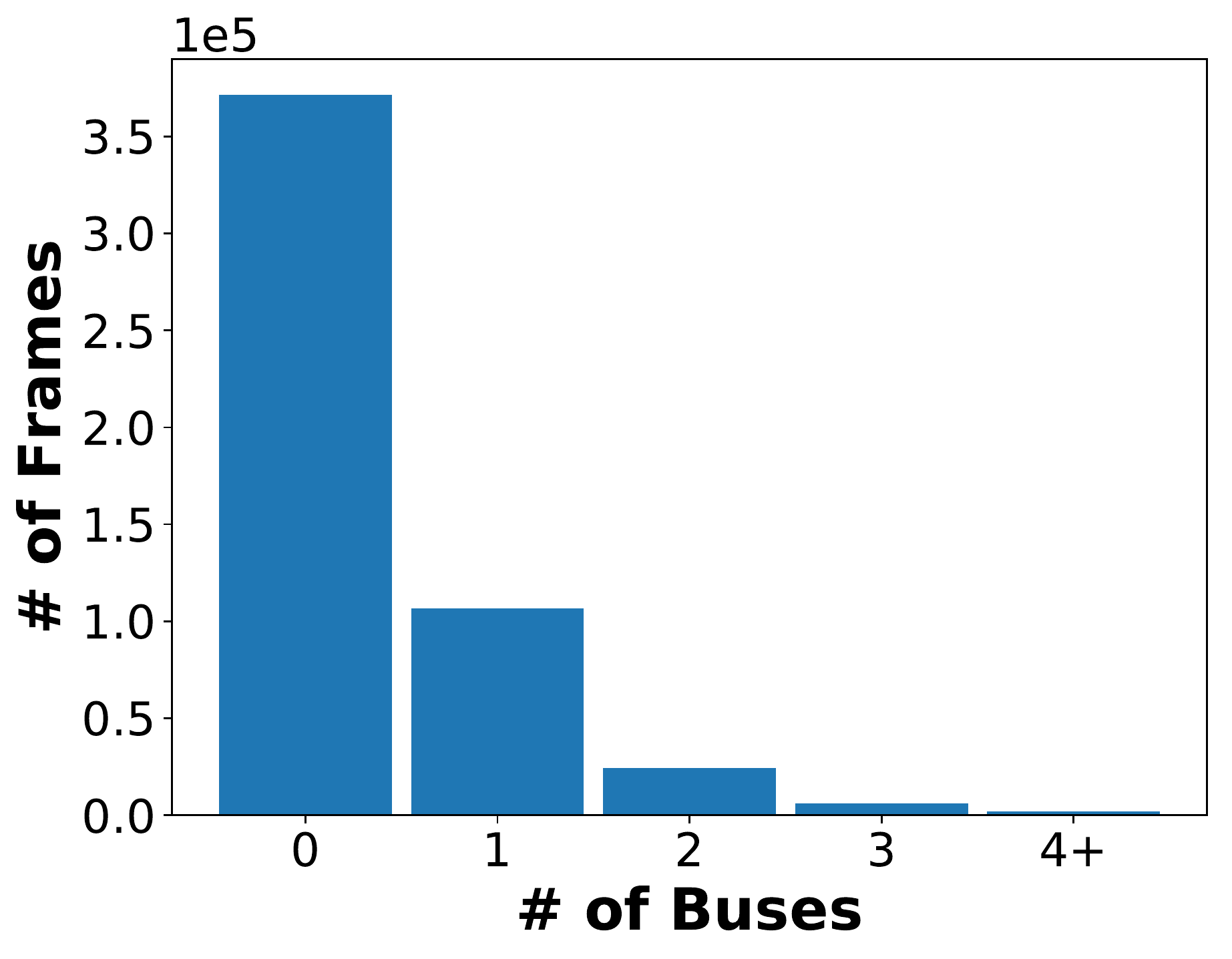}  \hfill $\ $ \\ \vspace{0.1in}
        \hfill
        \includegraphics[width=0.47\linewidth]{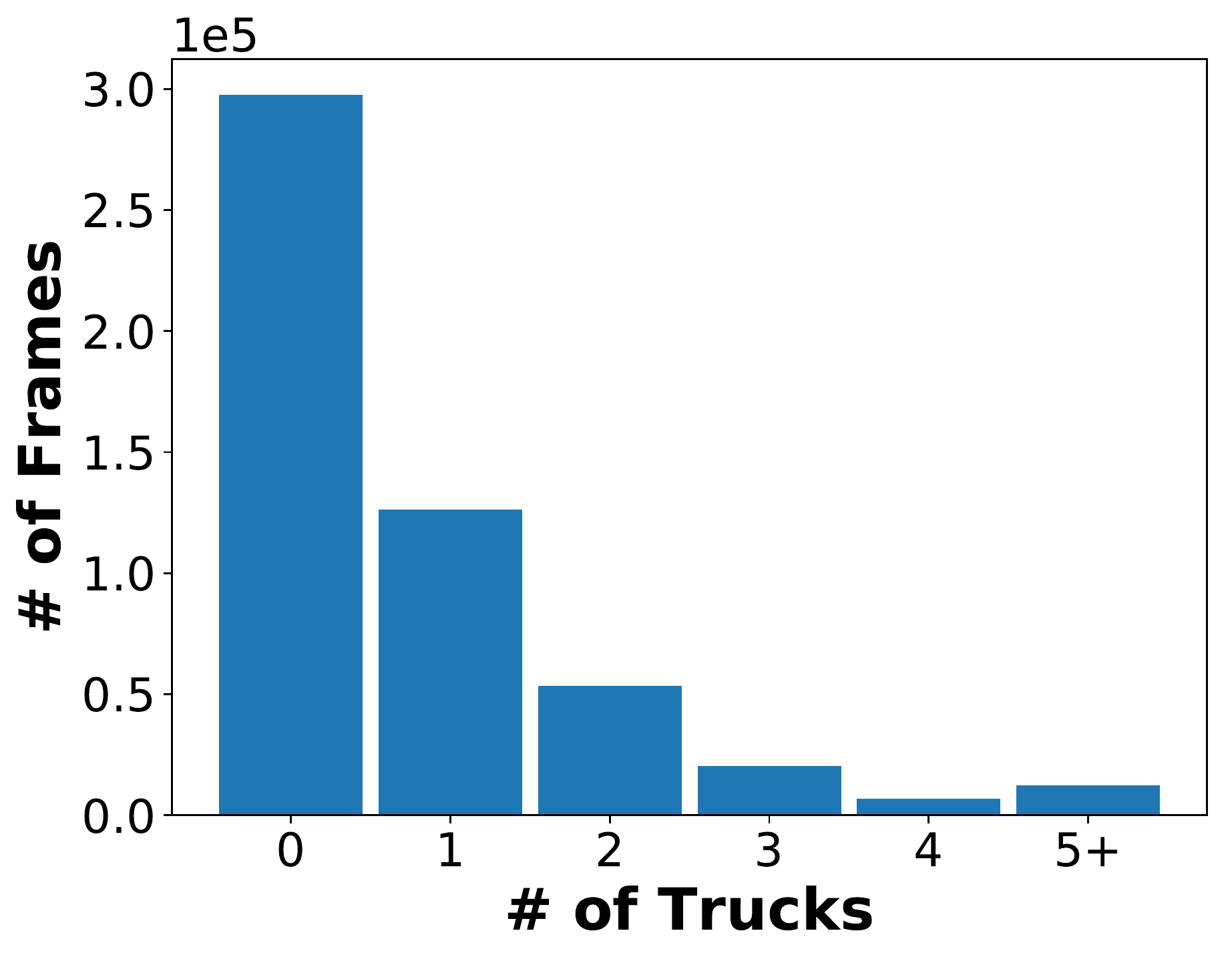} \hfill
        \includegraphics[width=0.47\linewidth]{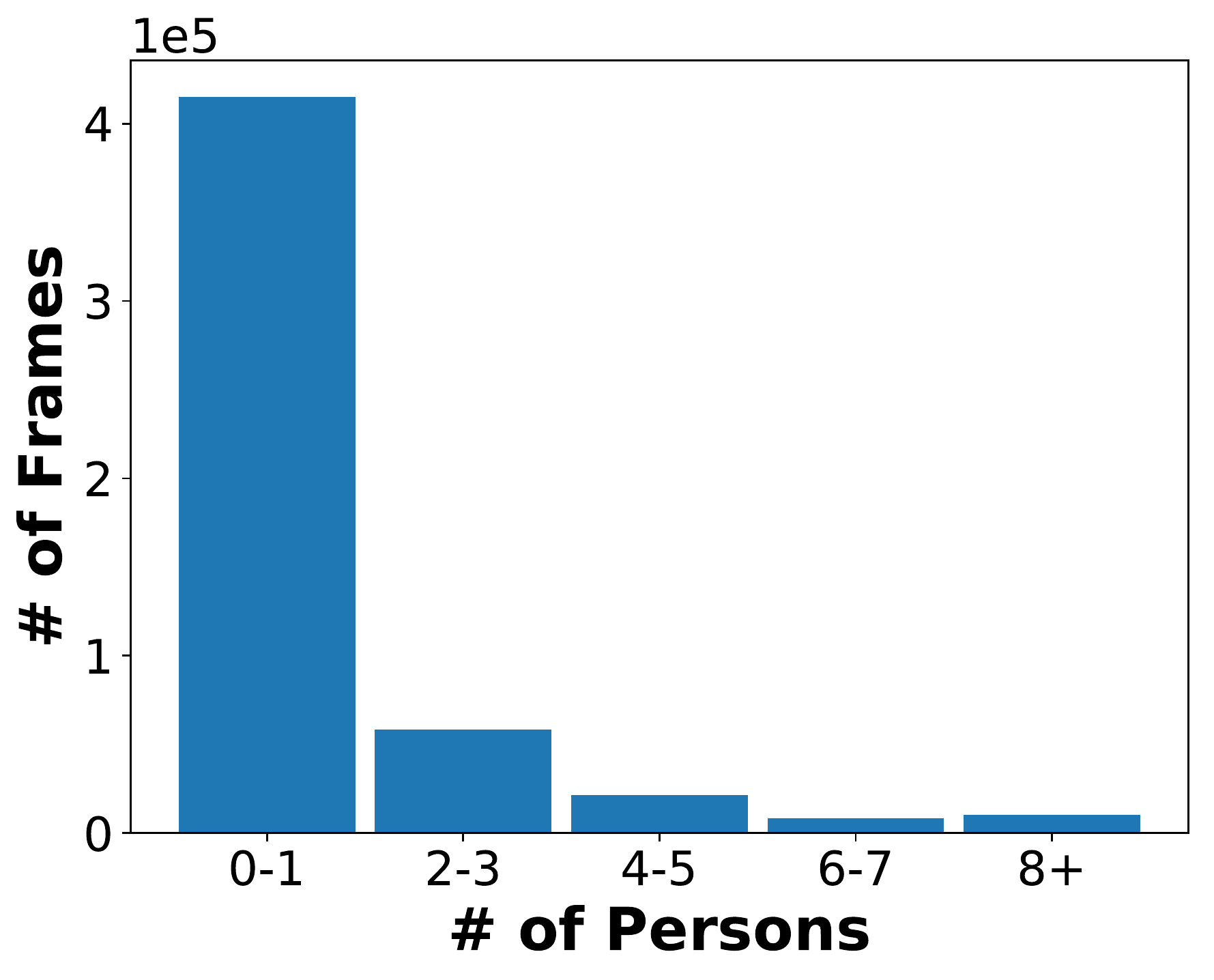}  \hfill $\ $
    \end{center}
    \caption{Statistics on the numbers of detection bounding boxes of 5 classes (\textit{car}, \textit{van}, \textit{bus}, \textit{truck}, and \textit{person}) in each frame.}
    \label{fig:obj-box}
\end{figure}

\begin{figure*}[htb]
    \begin{center}
        \hfill
        \includegraphics[width=0.23\linewidth]{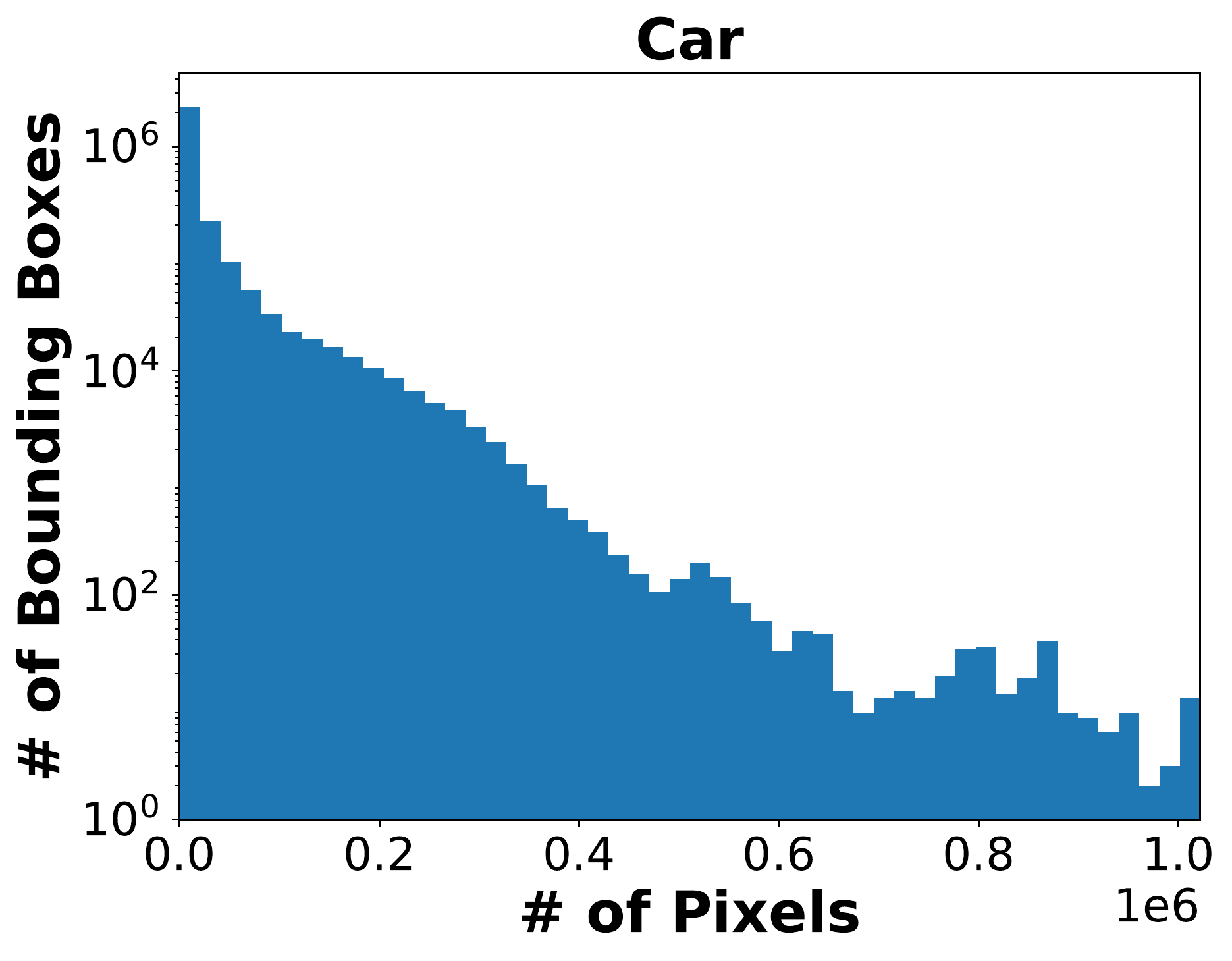}\hfill
        \includegraphics[width=0.23\linewidth]{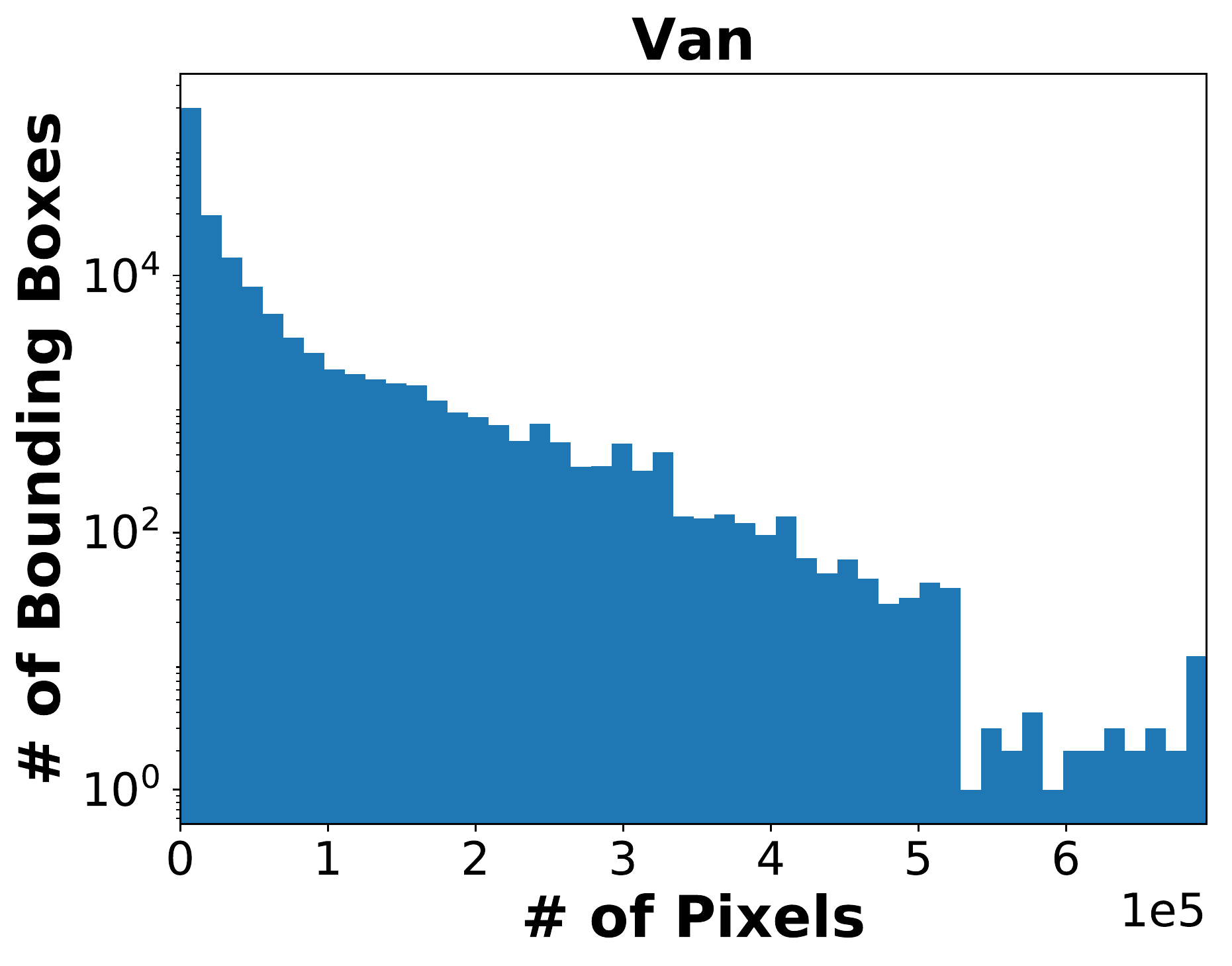}\hfill
        \includegraphics[width=0.23\linewidth]{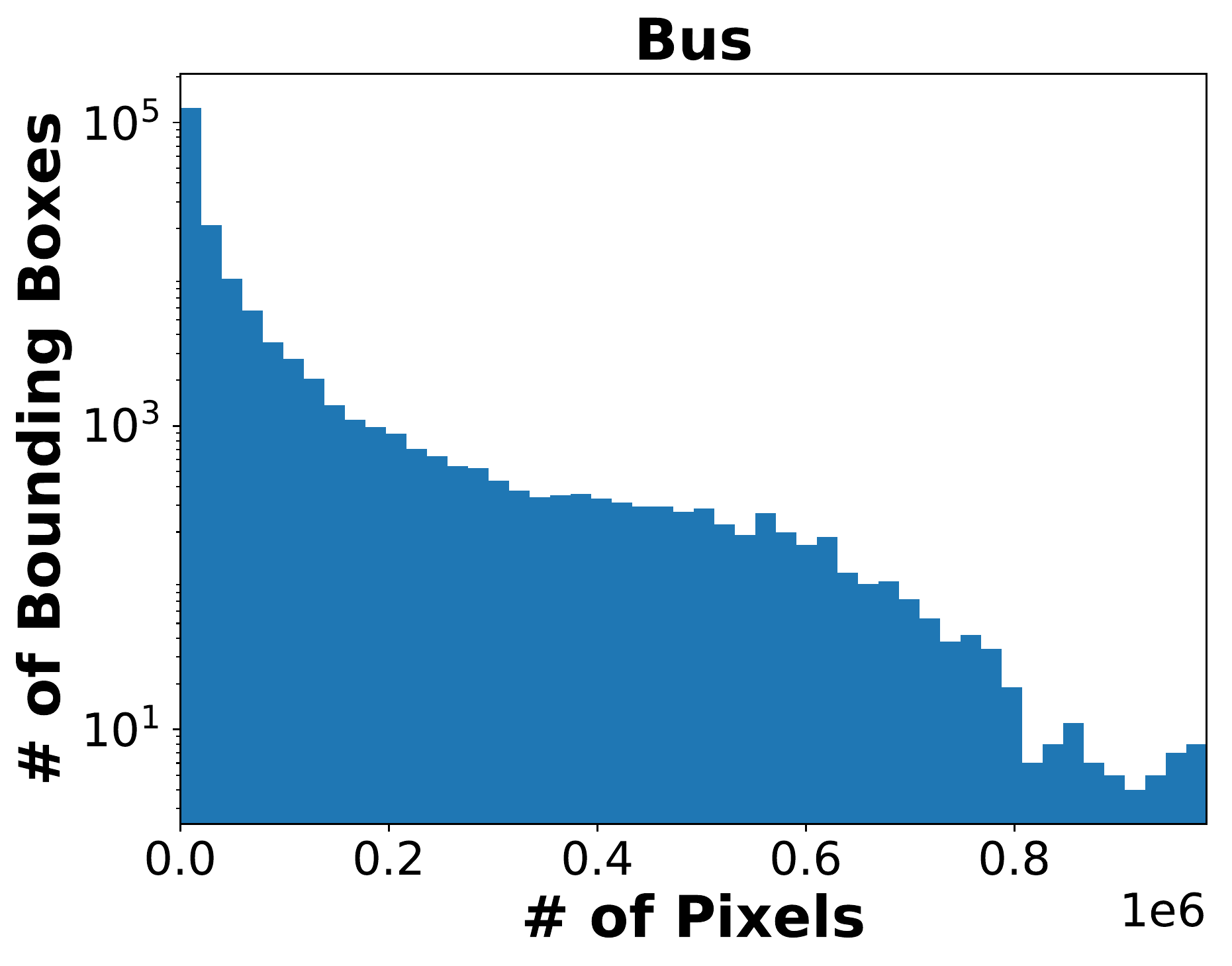}\hfill
        \includegraphics[width=0.23\linewidth]{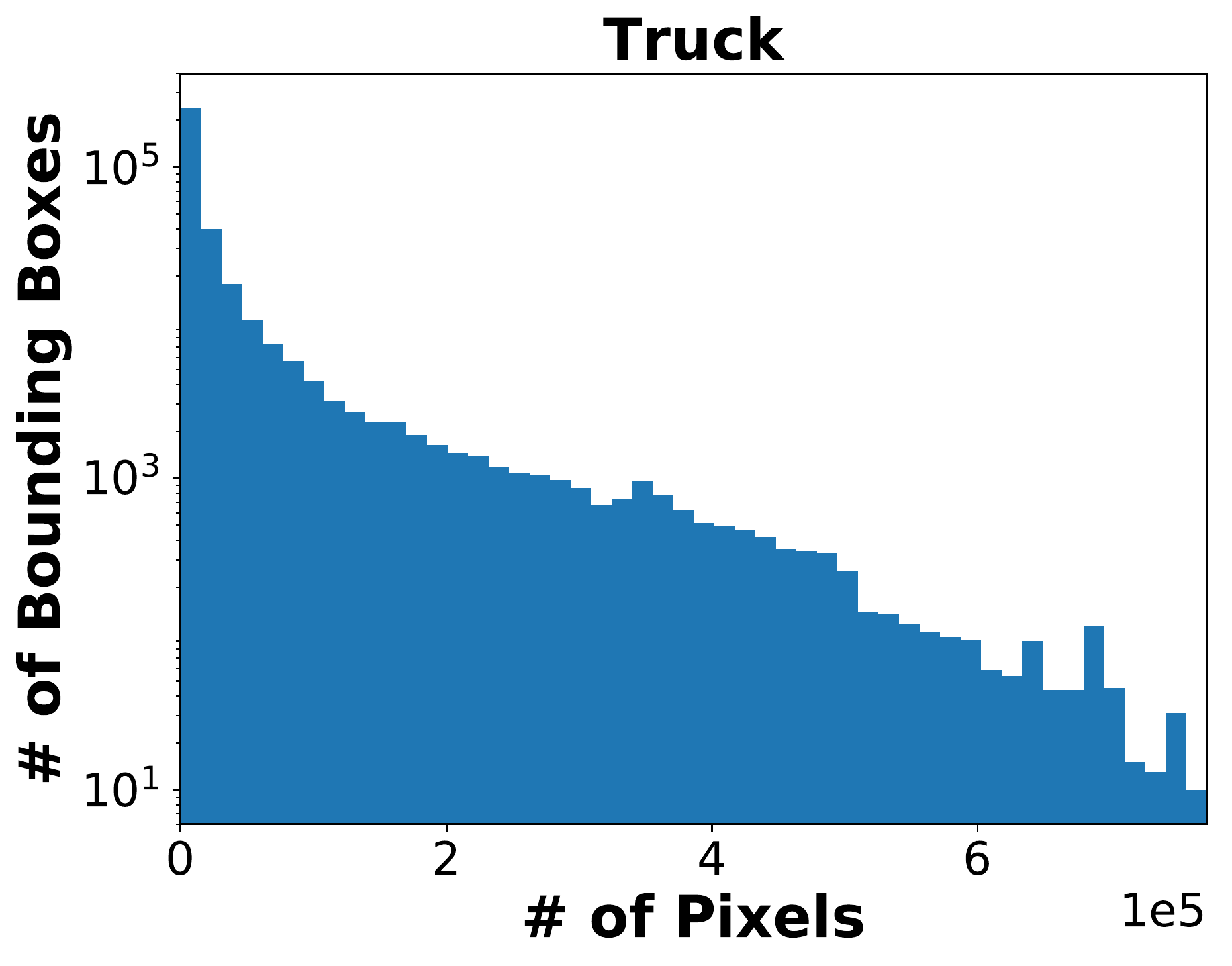}\hfill $\ $ \\ \vspace{0.1in}
        \hfill
        \includegraphics[width=0.23\linewidth]{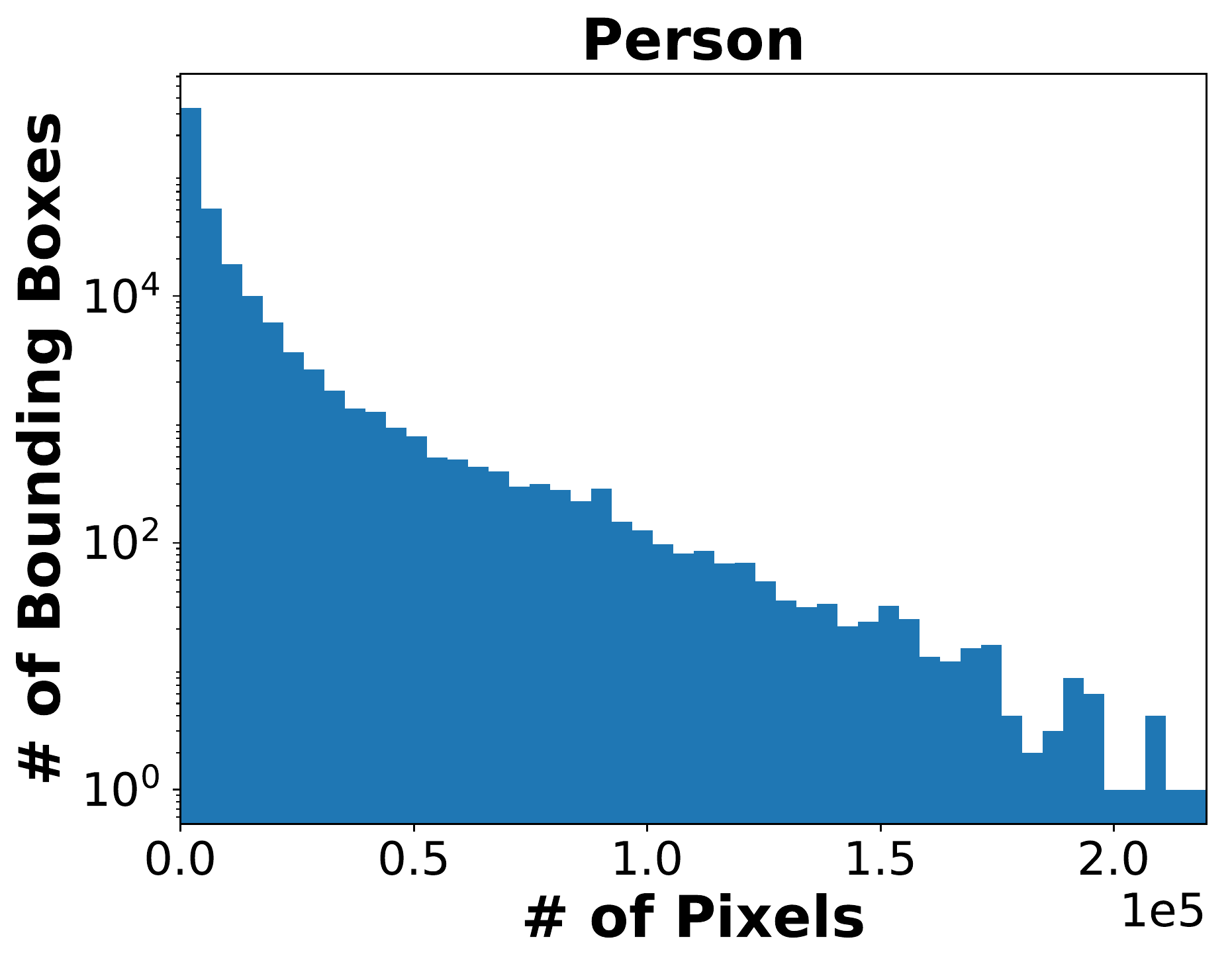}\hfill
        \includegraphics[width=0.23\linewidth]{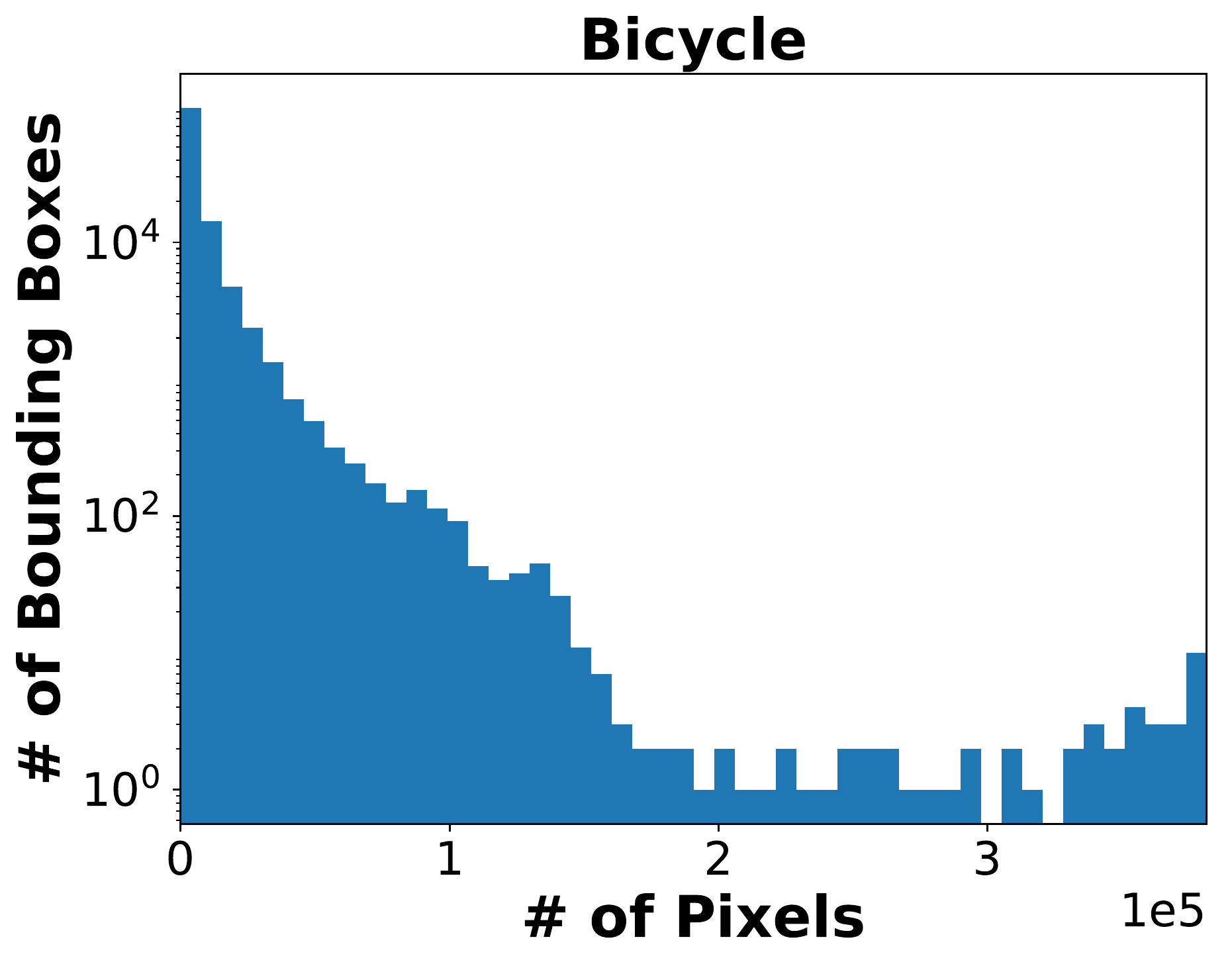}\hfill
        \includegraphics[width=0.23\linewidth]{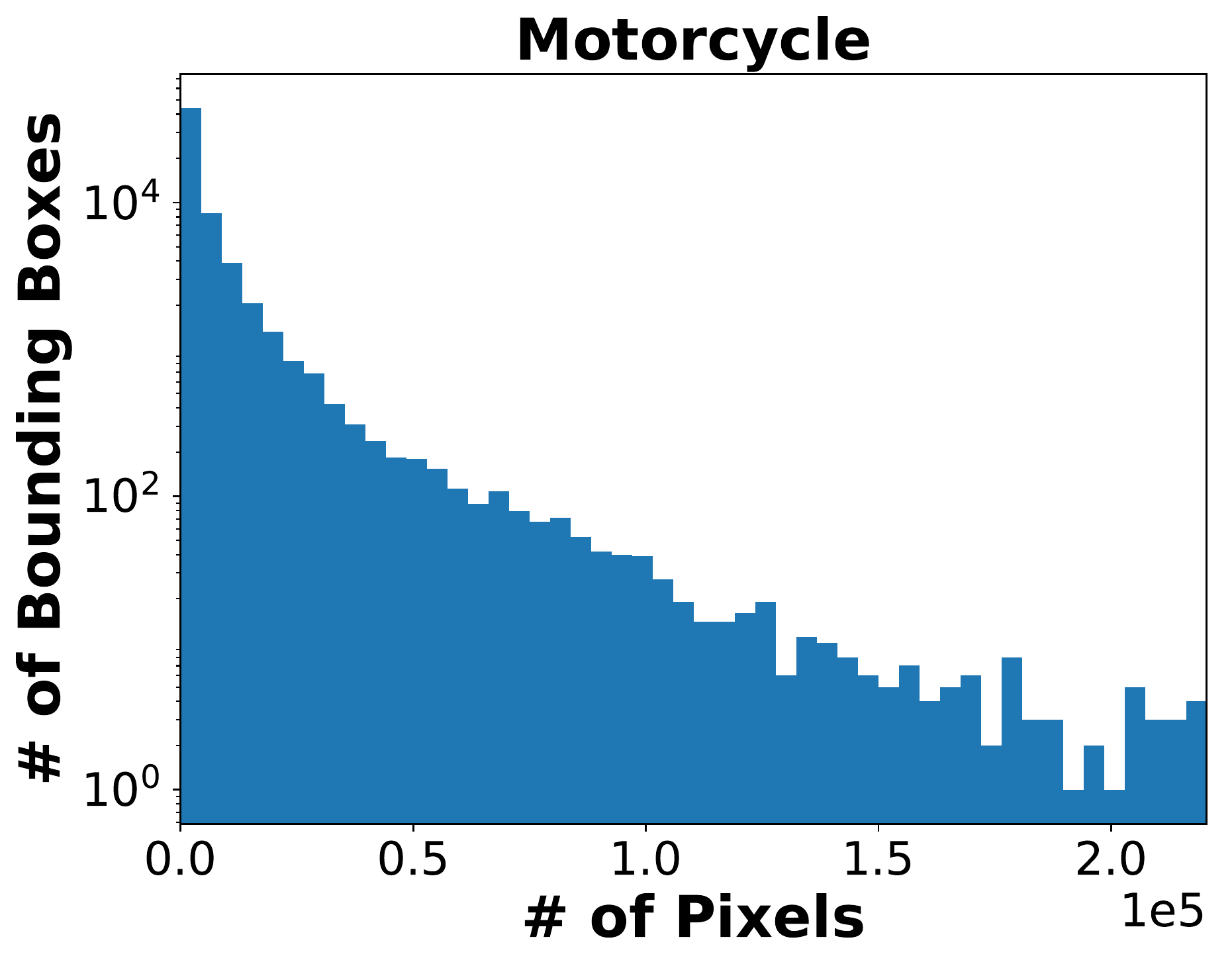}\hfill
        \includegraphics[width=0.23\linewidth]{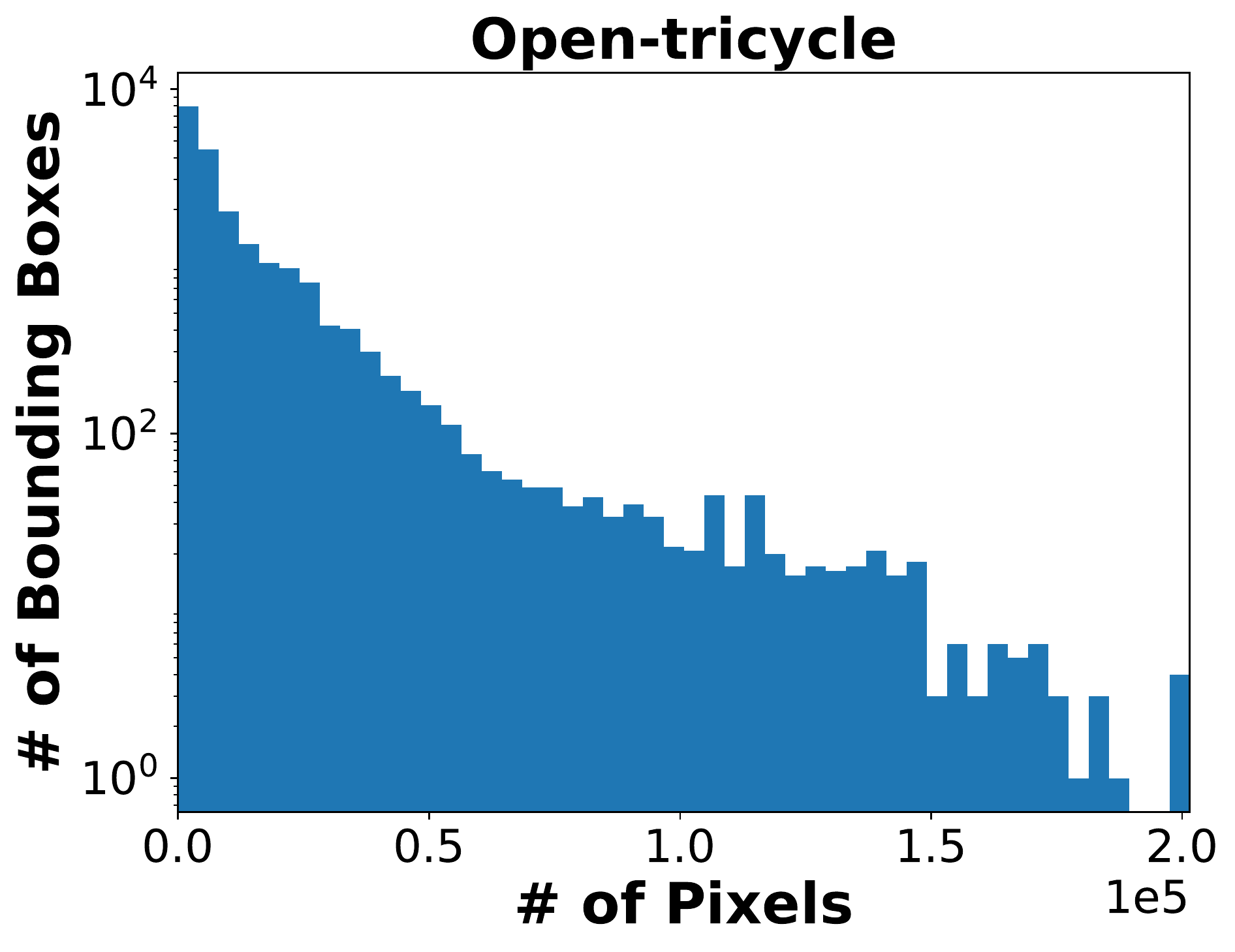}\hfill $\ $ \\ \vspace{0.1in}
        \hfill
        \includegraphics[width=0.23\linewidth]{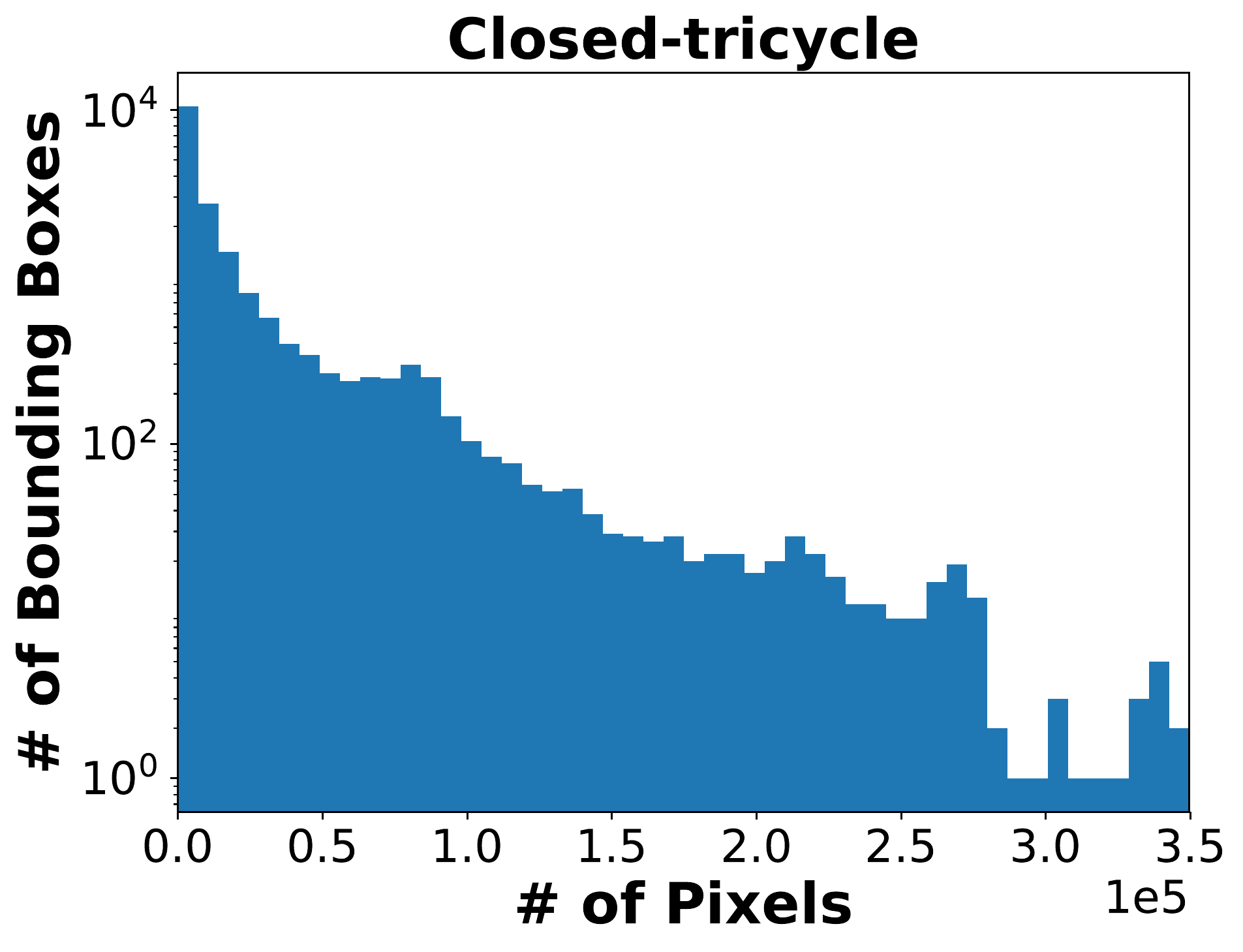}\hfill
        \includegraphics[width=0.23\linewidth]{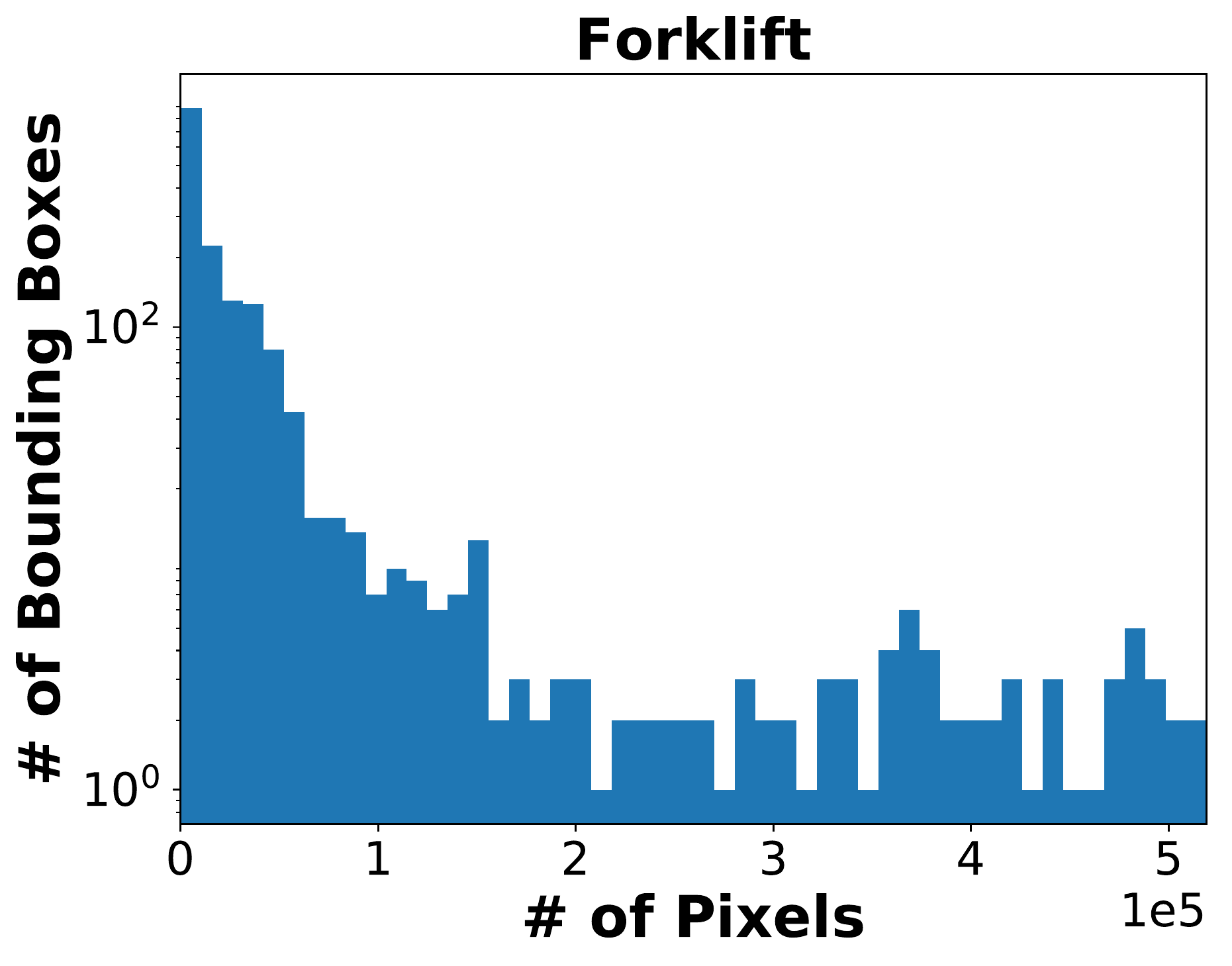}\hfill
        \includegraphics[width=0.23\linewidth]{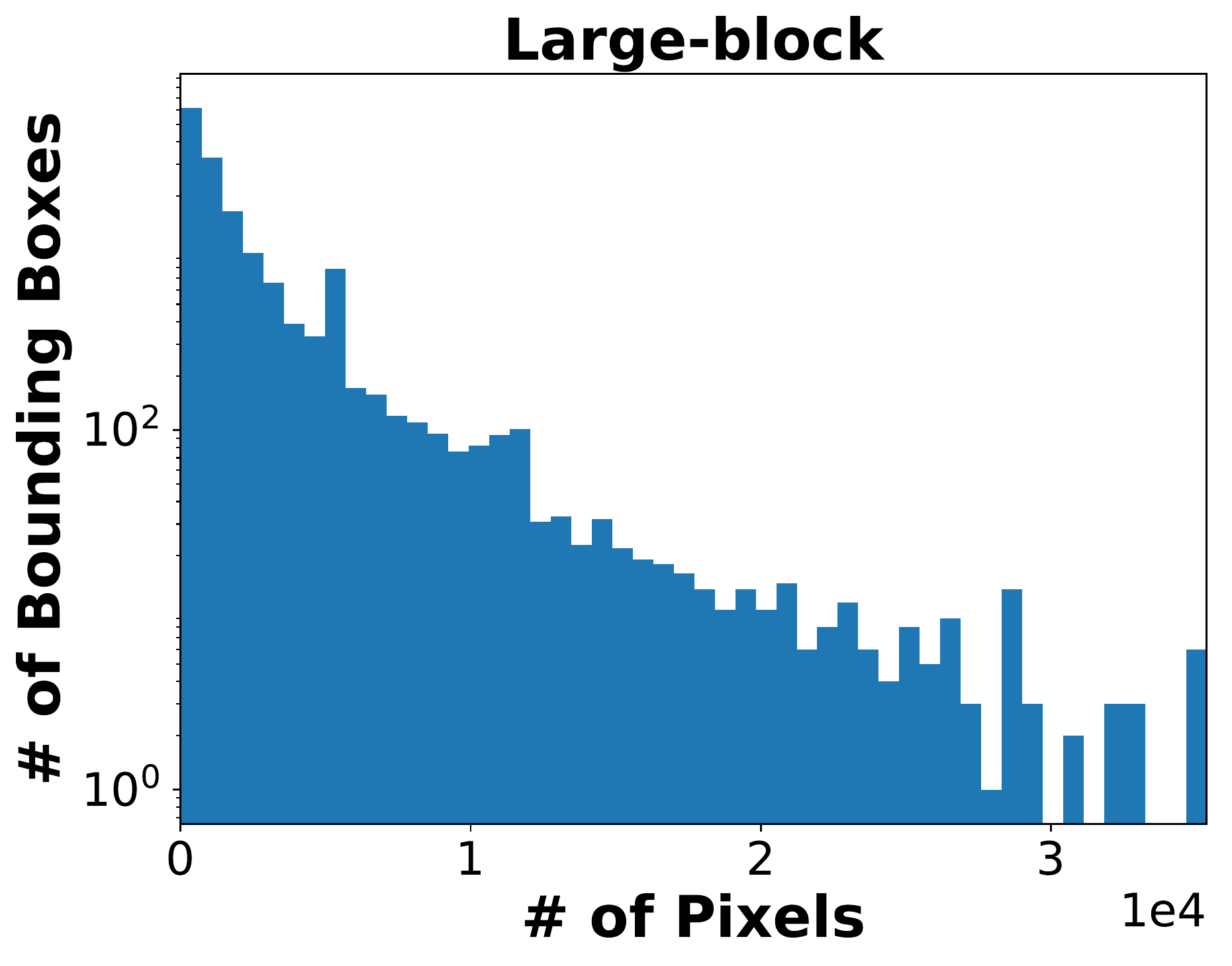}\hfill
        \includegraphics[width=0.23\linewidth]{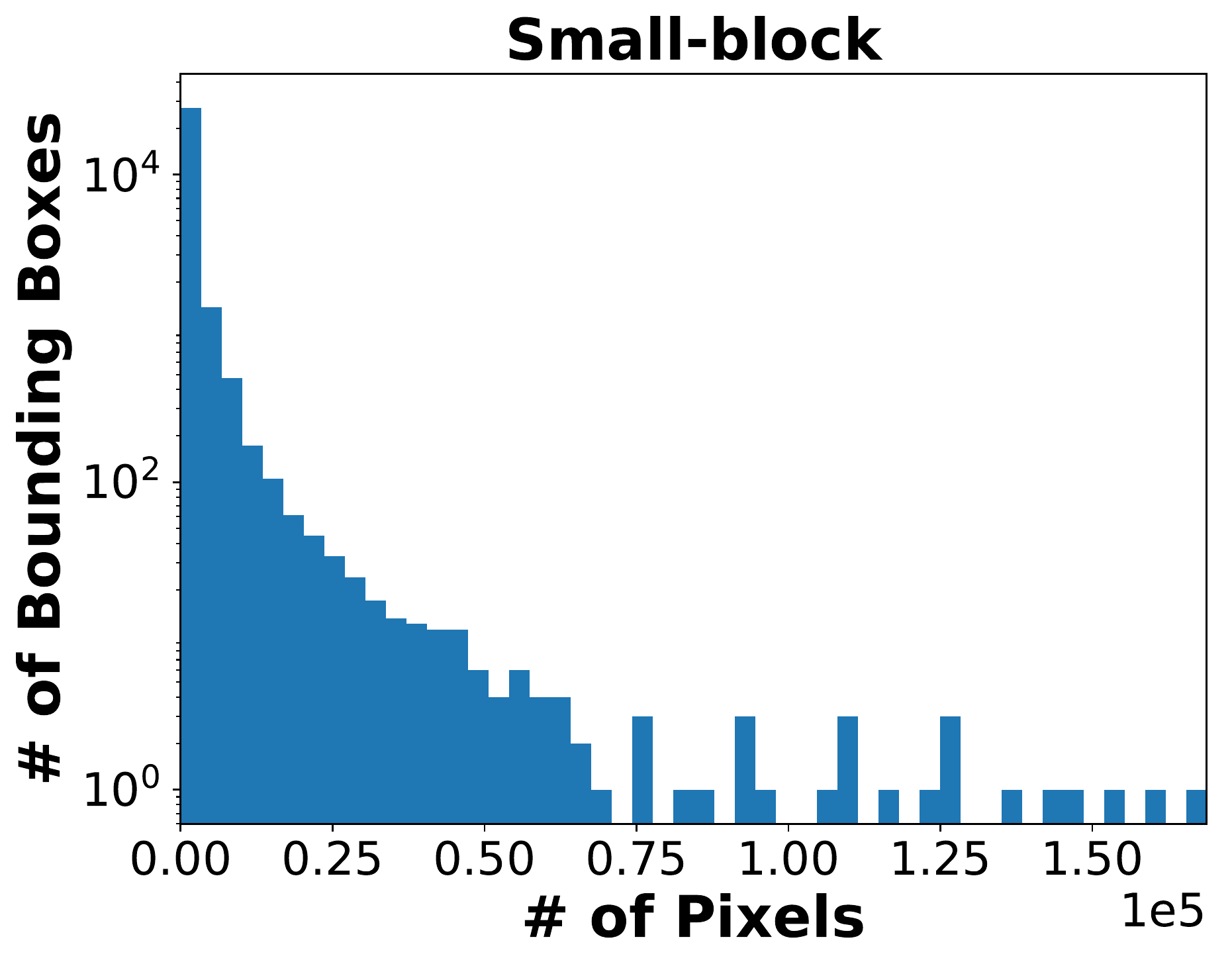}\hfill $\ $
    \end{center}
    \caption{Statistics of the number of pixels of bounding boxes of 12 classes.}
    \label{fig:obj-area}
\end{figure*}

\paragraph{Detection annotations}
Figure~\ref{fig:obj-box} shows the distributions of the numbers of objects for the 5 most common classes. As the two classes with most instances, there are on average $5.37$ cars and $0.85$ persons in each frame.
As shown in Table~\ref{tab:occluded-cut}, $45.23\%$ of the annotated objects are occluded by other objects in some degrees, and $5.71\%$ of the bounding boxes do not cover the full extent of the objects because the objects extend outside the image.
Table~\ref{tab:obj-area} shows the mean and median numbers of pixels of bounding boxes for the 12 classes in videos of 720p and 1080p separately.
The distributions of the numbers of bounding box pixels for the 12 classes are shown in Figure~\ref{fig:obj-area}.
Since there are two video resolutions (720p and 1080p), we scaled videos in 720p to 1080p and then calculated the bounding box areas.

\begin{figure}[b!]
    \begin{center}
        \hfill
        \includegraphics[width=0.47\linewidth]{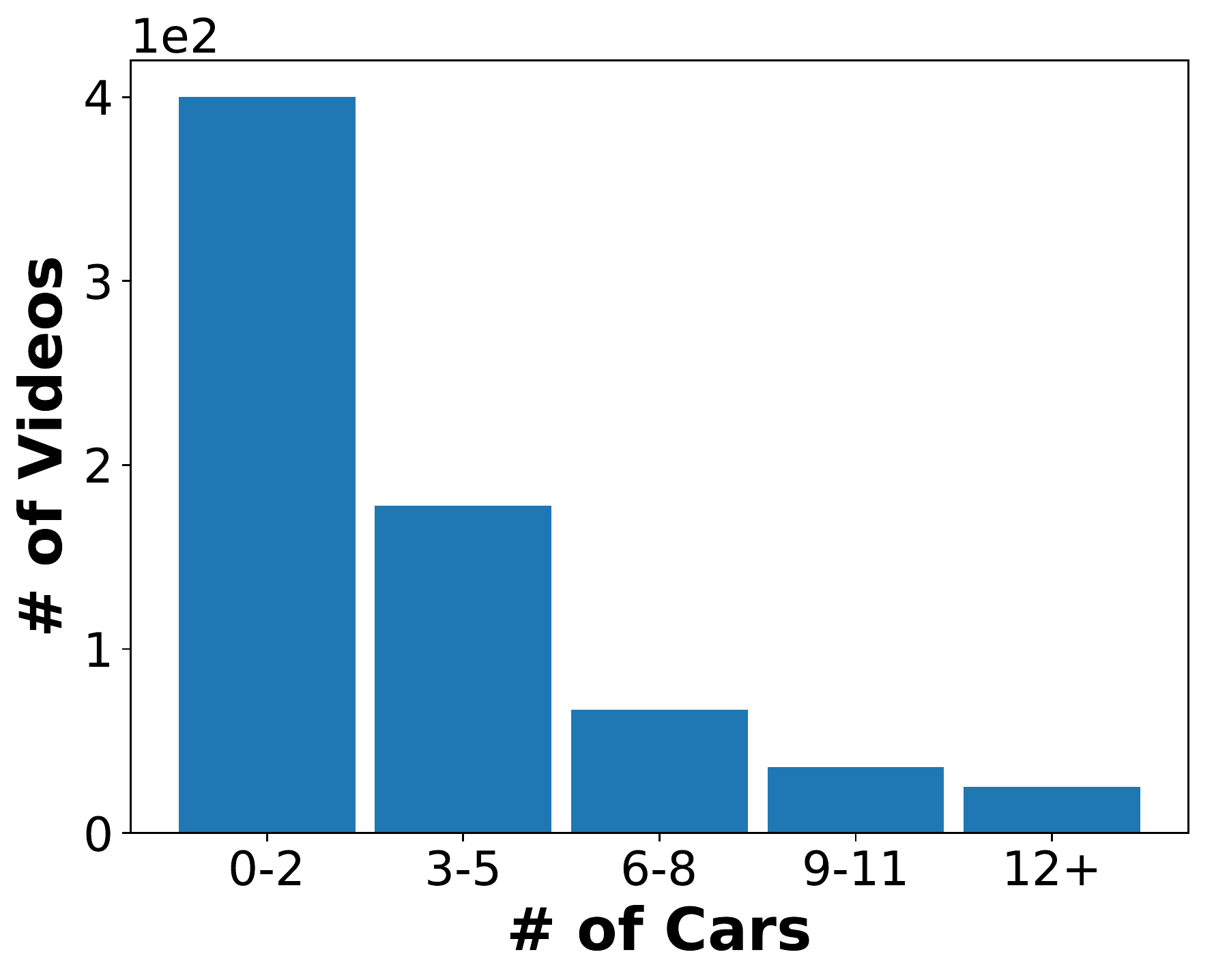} \hfill $\ $ \\ \vspace{0.1in}
        \hfill
        \includegraphics[width=0.47\linewidth]{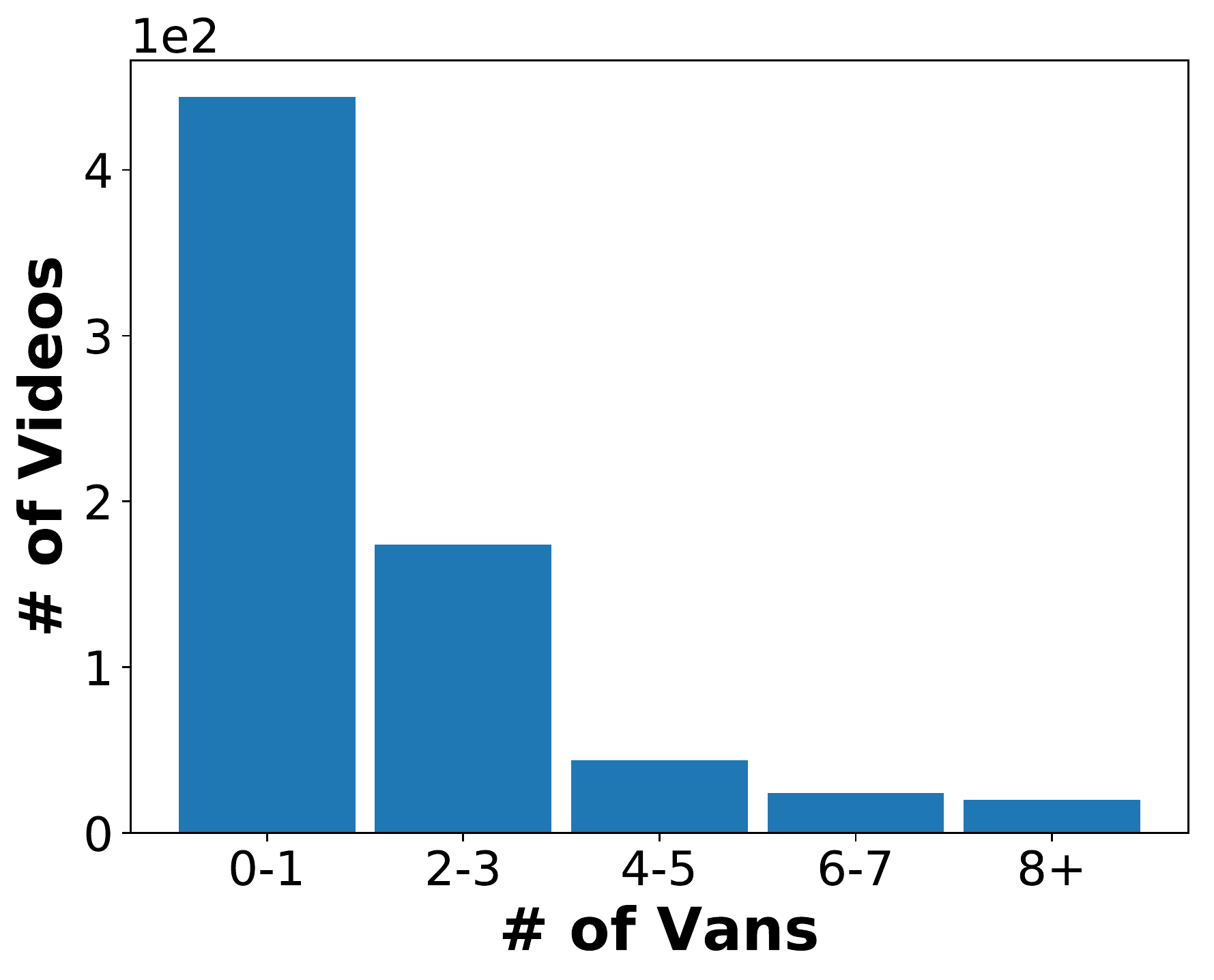} \hfill
        \includegraphics[width=0.47\linewidth]{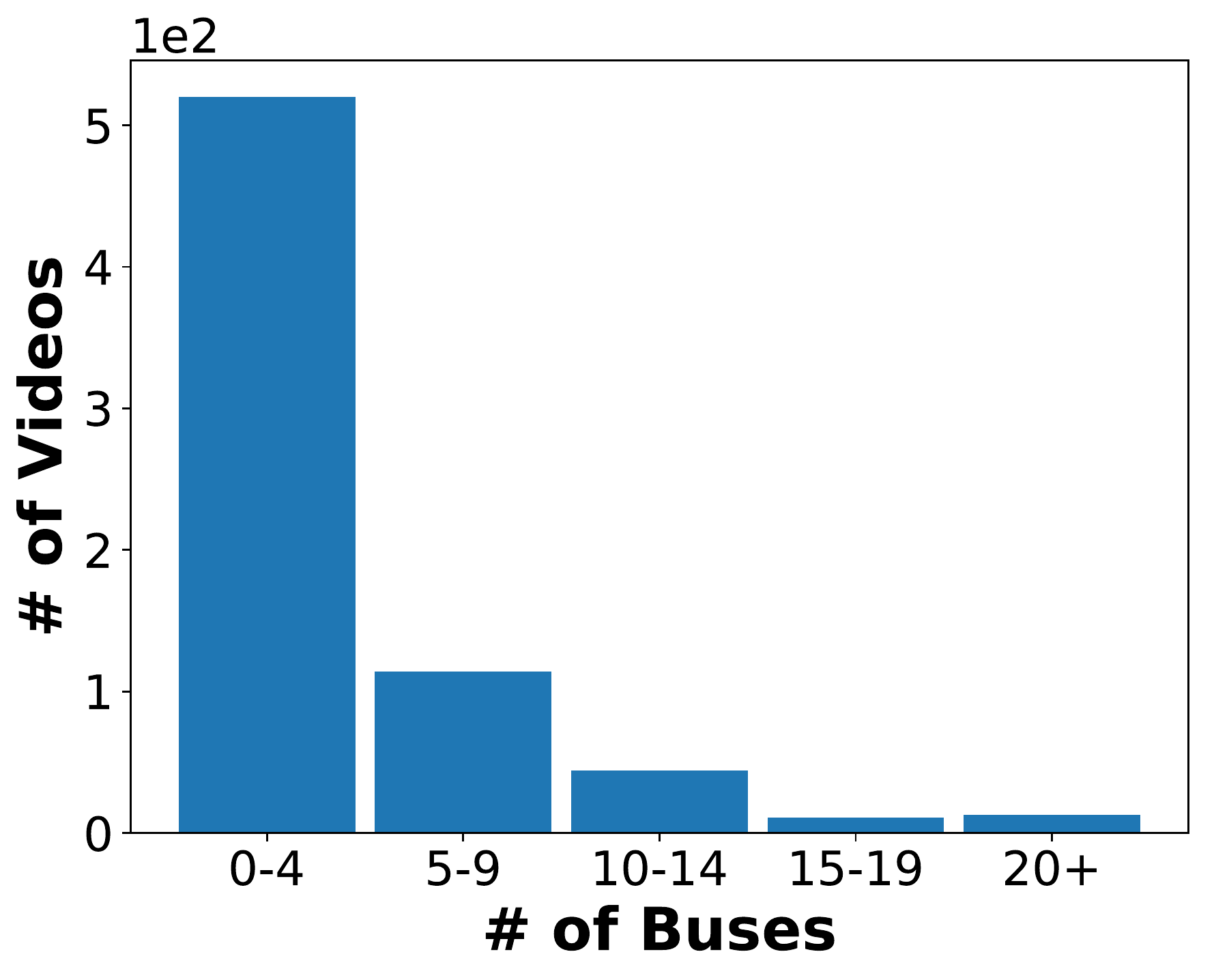} \hfill $\ $ \\ \vspace{0.1in}
        \hfill
        \includegraphics[width=0.47\linewidth]{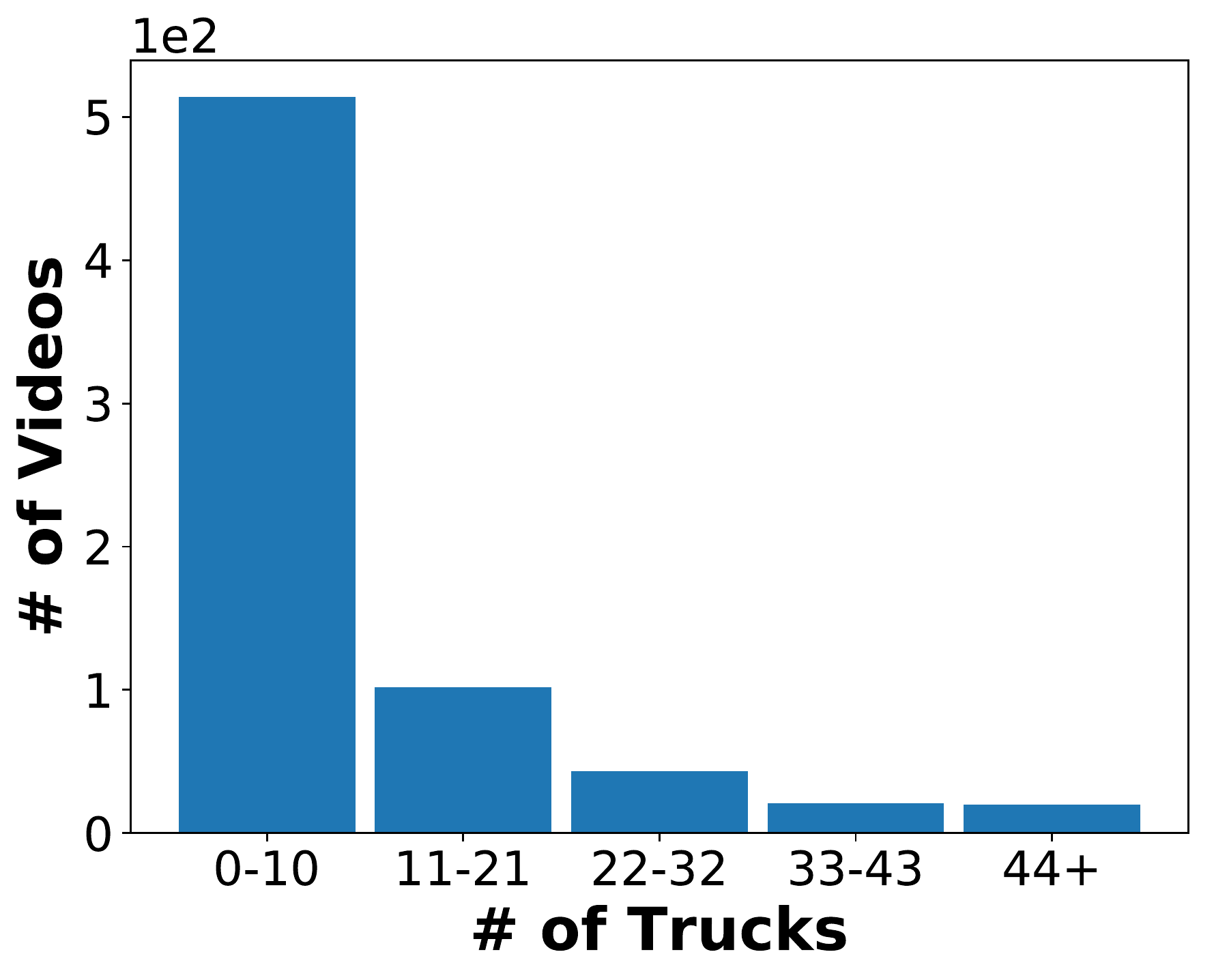} \hfill
        \includegraphics[width=0.47\linewidth]{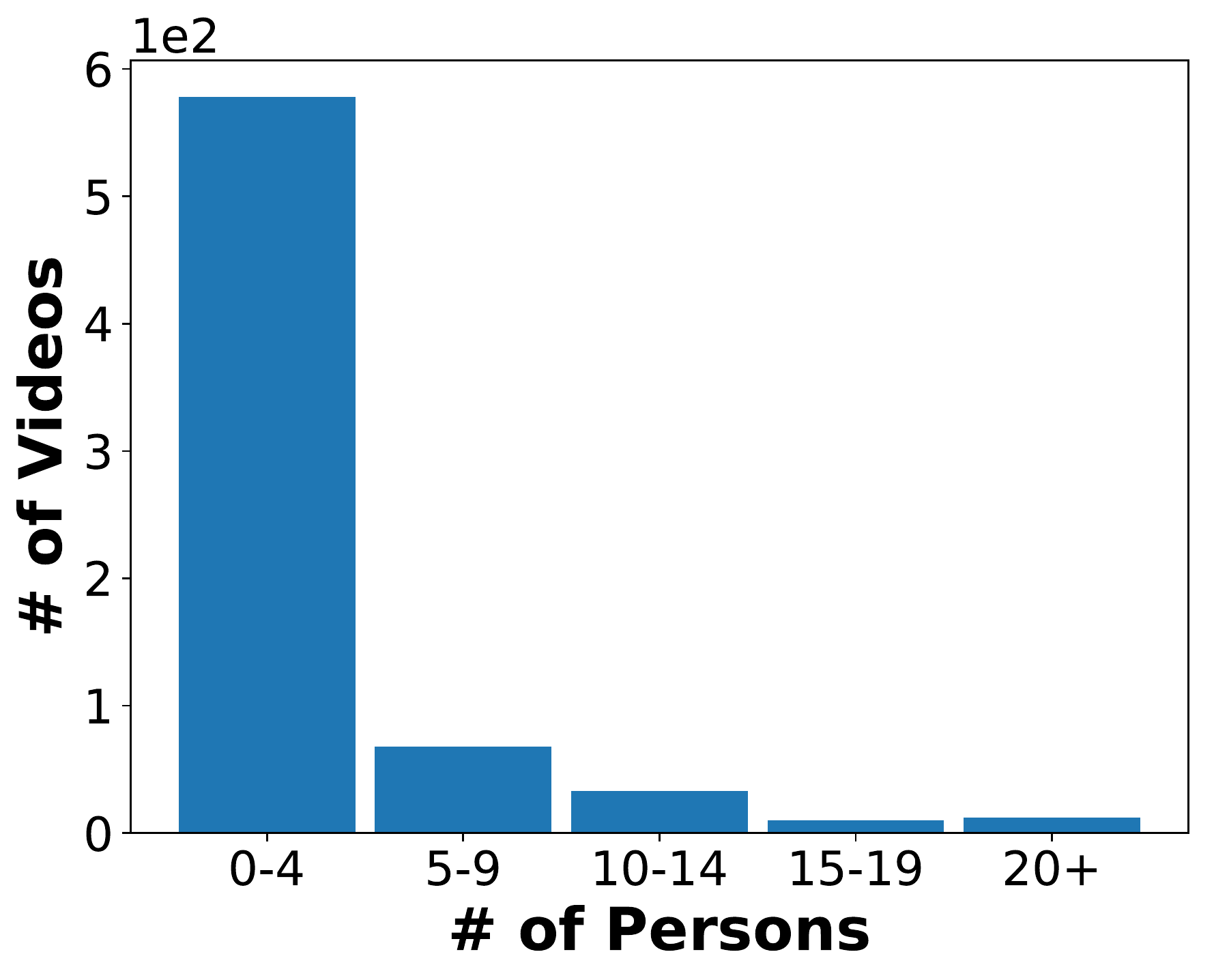} \hfill  $\ $
    \end{center}
    \caption{Statistics on the numbers of tracking instances of 5 classes (\textit{car}, \textit{van}, \textit{bus}, \textit{truck}, and \textit{person}) in each video.}
    \label{fig:obj-tracking}
\end{figure}

\paragraph{Tracking annotations}
Figure~\ref{fig:obj-tracking} shows the distributions of the numbers of tracked instances in each video for the 5 most common classes. There are on average $33.48$ cars and $8.46$ persons in each video.

\begin{table}[t]
    \begin{center}
        \begin{tabular}{c|c@{}c|c@{}c}
            \toprule
            & \multicolumn{2}{c|}{Occlusion (\textit{occluded})} & \multicolumn{2}{c}{Truncation (\textit{cut})} \\ \midrule
            \textit{no} & $2.33 \times {10}^{7}$ & $(54.77\%)$ & $4.02 \times {10}^{7}$ & $(94.29\%)$ \\ \cmidrule{1-5}
            \textit{partially} & $1.21 \times {10}^{7}$ & $(28.47\%)$ & $0.14 \times {10}^{7}$ & $(3.23\%)$ \\ \cmidrule{1-5}
            \textit{almost} & $0.71 \times {10}^{7}$ & $(16.76\%)$ & $0.11 \times {10}^{7}$ & $(2.48\%)$ \\
            \bottomrule
        \end{tabular}
    \end{center}
    \caption{Number of bounding boxes with different levels of occlusion (\textit{occluded}) and truncation (\textit{cut}) in the training set.}
    \label{tab:occluded-cut}
\end{table}

\begin{table*}[htb]
    \begin{center}
        \begin{tabular}{c|c|c|c|c|c|c}
            \toprule
            & \textit{car} & \textit{van} & \textit{bus}  & \textit{truck} & \textit{person} & \textit{bicycle} \\ \cmidrule{1-7}
            720p HD & $8300$ / $1750$ & $9320$ / $2091$ & $13249$ / $2714$ & $16280$ / $3400$ & $2538$ / $880$ & $3356$ / $1386$  \\ \cmidrule{1-7}
            1080p FHD & $17583$ / $3600$ & $22483$ / $4508$ & $42488$ / $8436$ & $32869$ / $6552$ & $4011$ / $1598$ & $5604$ / $2255$  \\ \midrule
            & \textit{motorcycle} & \textit{open-tricycle} & \textit{closed-tricycle}  & \textit{forklift} & \textit{large-block} & \textit{small-block}\\ \cmidrule{1-7}
            720p HD & $3114$ / $1122$ & $5118$ / $2255$ & $16942$ / $6440$ & $12008$ / $3300$ & $1147$ / $555$ & $525$ / $231$  \\ \cmidrule{1-7}
            1080p FHD & $5989$ / $1887$ & $14486$ / $6570$ & $11990$ / $4218$ & $44936$ / $9828$ & $2077$ / $870$ & $1680$ / $561$  \\
            \bottomrule
        \end{tabular}
    \end{center}
    \caption{Mean/Median of the numbers of pixels in one bounding box for 12 classes.}
    \label{tab:obj-area}
\end{table*}

\section{Related Work}
Recent revolutions in computer vision applications are partly credited to the increasing number of publicly available datasets with dense annotations, as these datasets have provided massive data with detailed information, novel challenges, and fair benchmarks to this community.
Large generic vision datasets and challenges of a broader set of objects and scenes, such as PASCAL VOC~\cite{everingham2010pascal},
ImageNet~\cite{deng2009imagenet},  COCO~\cite{lin2014microsoft}, and Open Images~\cite{kuznetsova2018open} have pushed forward the frontiers of image classification, object detection, segmentation, and even learning better hierarchical representations from data.
The advances in developing general visual recognition and perception abilities also encouraged the research and applications in specific tasks and domains with complex challenges, unseen situations, promising prospects, and great demands of large-scale high-quality datasets, such as intelligent driving.

Traffic datasets in earlier stages were mainly for pedestrian detection and tracking~\cite{dollar2009pedestrian,enzweiler2009monocular,ess2008mobile,wojek2009multi}.
Due to limits on data scale and scope, these datasets can not fully represent various traffic scenes.
Other datasets focused on semantic and instance segmentations for scene understanding.
CamVid~\cite{brostow2008segmentation} was one of the first driving datasets with pixel-level semantic annotations.
CityScapes~\cite{cordts2016cityscapes} provided $5000$ and $20,000$ images with fine and coarse annotations, respectively, for 30 classes.
The Mapillary Vistas Dataset~\cite{neuhold2017mapillary} provided even more semantic categories and instance-specific annotations on $25,000$ images.
However, these datasets only had annotations on frame level, and the expensive cost of per-pixel labeling made it unrealistic to create larger-scale datasets with segmentation annotations.
ApolloScape~\cite{huang2018apolloscape} provided large-scale datasets collected in China with main focus on 3D point cloud and 3D-2D segmentations. For 2D videos, it provided $140,000$ frames with pixel-level annotations and selected about $6000$ images as semantic segmentation benchmark from a relatively small set of videos.

Detection and tracking-by-detection is another research direction for scene understanding, and bounding box labeling is much more convenient than pixel-level one.
The TME Motorway dataset~\cite{caraffi2012system} provided 28 videos clips with more than $30,000$ frames along with bounding box tracking annotations from laser-scanner.
The KITTI Vision Benchmark Suite~\cite{geiger2013vision} proposed several tasks as benchmarks and provided data from different sensors.
However, these two datasets lacked of diversity and complexity in scenes, as they only collected a limited set of data.
The nuScenes Dataset~\cite{nuscenes2019} aimed at providing data from the entire sensor suite, similar to KITTI, rather than only camera-based vision.
It was seven times larger than KITTI in object annotations.
However, it only annotated $40,000$ keyframes from its front-facing cameras among all $1.4$ million images from all 6 cameras, and the provided annotations are more beneficial to detection rather than tracking task.
BDD100K~\cite{yu2018bdd100k} and the earlier NEXET~\cite{nexar2017nexar} dataset collected driving videos from a single face-front camera in a crowdsourcing way.
While BDD100K released a large collection of $100,000$ raw videos, only one frame from each video was annotated.

Compared with existing datasets, {\dc} provided more than $10,000$ driving videos from hundreds of drivers. It better covered different traffic conditions and driving behaviors and fully respected the diversity and complexity in real-world traffic scenes.
Our densely annotated detection and tracking information on $1000$ videos made the large video collection more helpful to the field of intelligent driving.

\section{Conclusion and Future Work}
In this work, we propose a large-scale driving dataset named {\dc}, which contains more than $10,000$ raw videos and detailed bounding box and tracking annotations for $1000$ videos within the current release.
As one of the largest publicly available driving datasets providing both video sequences and detection and tracking annotations, {\dc} enriches the community with various and complex traffic scenarios, large-scale exhaustive annotations, and novel and challenging tasks.
In the future, we plan to release the remaining keyframe annotations, further enlarge the collections of data, and expand the coverage of all-weather scenarios, extreme conditions, and rare cases.

\paragraph{Acknowledgement}
We would thank Yi Yang, Yifei Zhang, and Guozhen Li and their teams for their support on data collection.
We would thank Haifeng Shen, Xuelei Zhang, and Wanxin Tian for their support on privacy protection.
We would thank Wensong Zhang, Yichong Lin, Runbo Hu, Yiping Meng, Menglin Gu, and many other people for useful discussions and general support.
We would thank drivers on DiDi's platform who consented to sharing dashcam videos.

{\small
\bibliographystyle{ieee}
\bibliography{reference}
}

\newpage
\appendix
\section{Labeling Descriptions and Class Examples}
\label{sec:class-def}
Table~\ref{tab:class-example} shows some object examples of each class to be annotated in the video frames.
We ignored all objects with no more than 25 pixels.

\begin{figure}[htb]
    \begin{center}
        \includegraphics[width=0.815\linewidth]{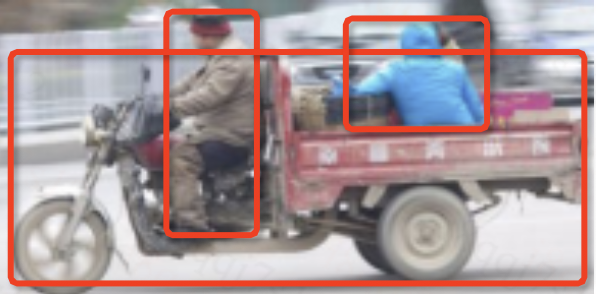}
    \end{center}
    \caption{Annotations of a group with three bounding boxes, including one \textit{open-tricycle} and two \textit{person} objects. We annotate them separately and assign the same \textit{group\_id} attributes to them.}
    \label{fig:example-group}
\end{figure}

\paragraph{{Cycles, riders and \textit{group\_id}}}
The bounding boxes for all cycles (bicycles, motorcycles, and tricycles) only cover the vehicles themselves, no matter riders and passengers are on the vehicles or not.
The driver, rider and passengers of bicycles, motorcycles, and open-tricycles are annotated as \textit{person} separately.
We additionally provide a \textit{group\_id} attribute for cycle vehicles. A cycle vehicle and its rider and passengers on it share the same \textit{group\_id} value.
In this way, we can easily group the corresponding people and vehicle together and treat them as a single instance of \textit{rider} class which is sometimes defined in other driving datasets~\cite{cordts2016cityscapes}. An example of a group with one open tricycle and two people is shown in Figure~\ref{fig:example-group}.
For drivers and passengers in closed-door tricycles and other closed vehicles, we do not annotate them explicitly.

\paragraph{Occlusions and truncations}
For each bounding box, we annotate the object is occluded (the \textit{occluded} attribute) by other objects or is out of the image frame boundary (the \textit{cut} attribute).
We do not label the bounding boxes of objects which are completely hidden behind other objects, even though it can be inferred by adjacent frames.

\begin{table*}
    \begin{center}
        \begin{tabular}{l|c|c}
            \toprule
            Class Name & Class ID & Examples in {\dc} \\
            \hline \hline
            \textit{car} & 1 &
                \includegraphics[height=.09\linewidth]{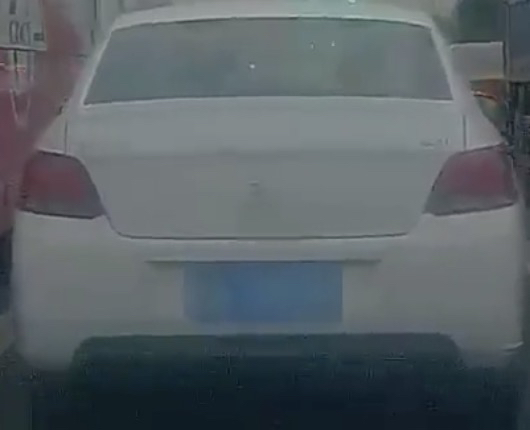} 
                \includegraphics[height=.09\linewidth]{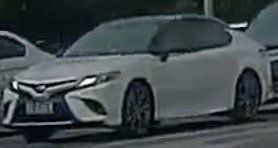} 
                \includegraphics[height=.09\linewidth]{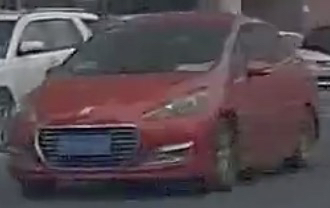}\\ \hline
            \textit{van} & 2 &
                \includegraphics[height=.09\linewidth]{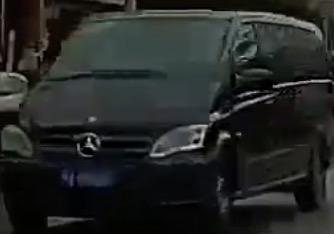} 
                \includegraphics[height=.09\linewidth]{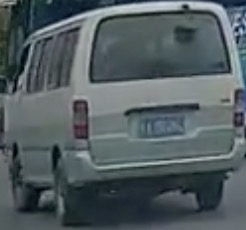} 
                \includegraphics[height=.09\linewidth]{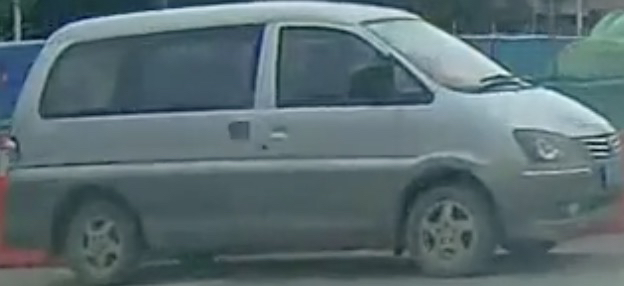}\\ \hline
            \textit{bus} & 3 &
                \includegraphics[height=.09\linewidth]{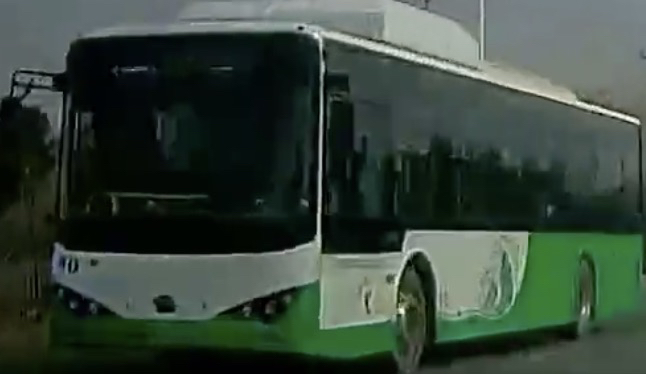} 
                \includegraphics[height=.09\linewidth]{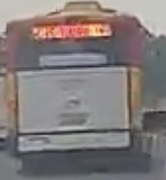} 
                \includegraphics[height=.09\linewidth]{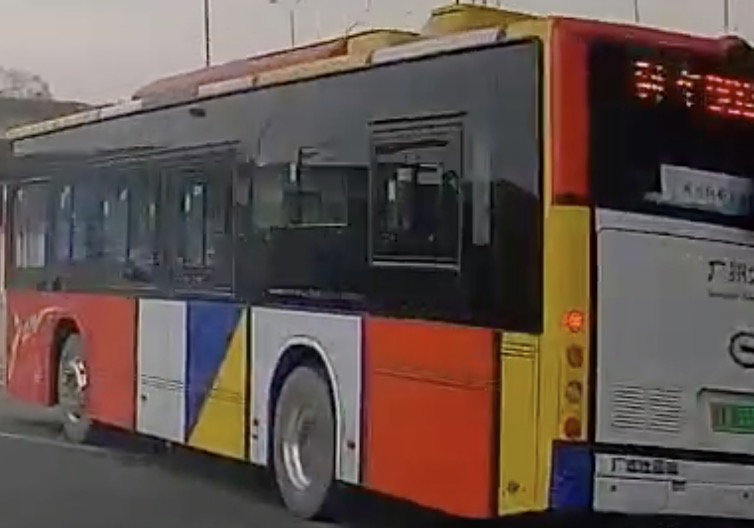}\\ \hline
            \textit{truck} & 4 &
                \includegraphics[height=.09\linewidth]{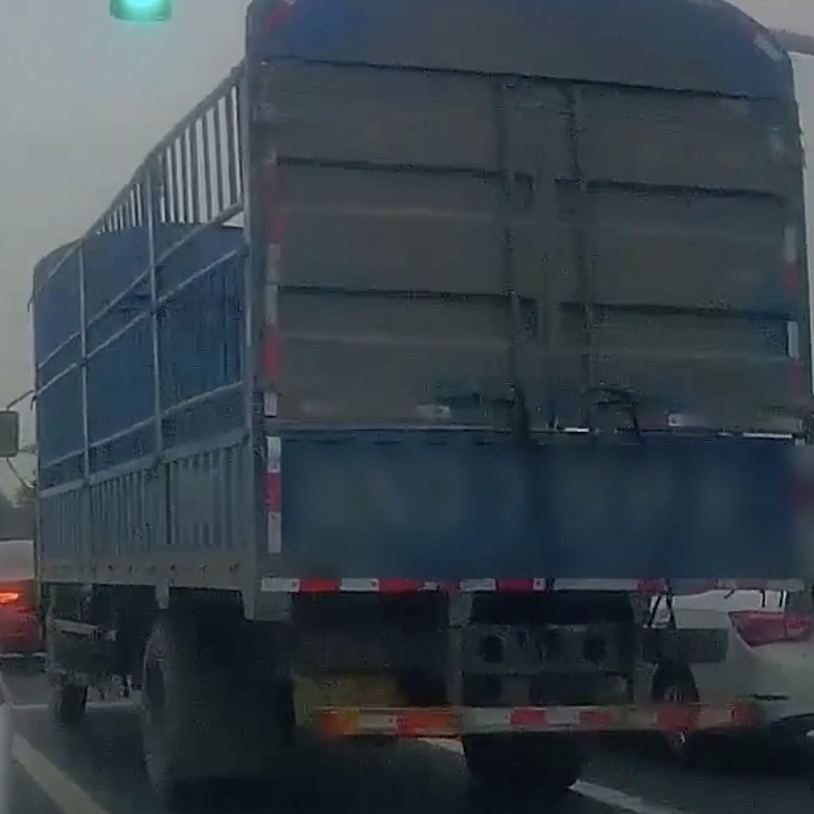} 
                \includegraphics[height=.09\linewidth]{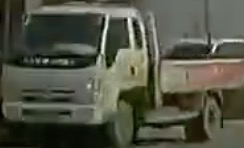} 
                \includegraphics[height=.09\linewidth]{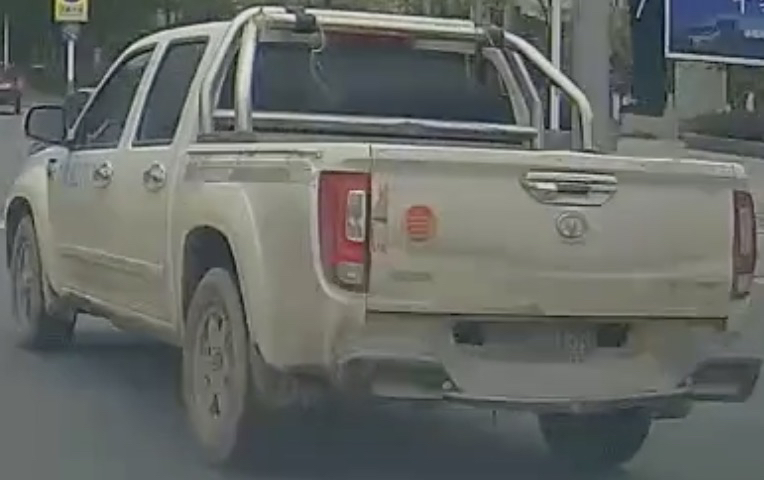}\\ \hline
            \textit{person} & 5 &
                \includegraphics[height=.09\linewidth]{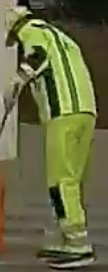} 
                \includegraphics[height=.09\linewidth]{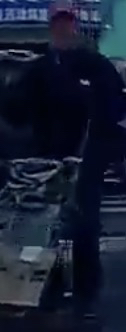} 
                \includegraphics[height=.09\linewidth]{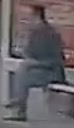}\\ \hline
            \textit{bicycle} & 6 &
                \includegraphics[height=.09\linewidth]{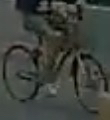} 
                \includegraphics[height=.09\linewidth]{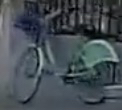} 
                \includegraphics[height=.09\linewidth]{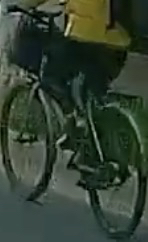}\\ \hline
            \textit{motorcycle} & 7 &
                \includegraphics[height=.09\linewidth]{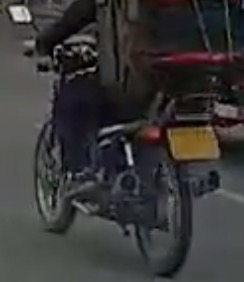} 
                \includegraphics[height=.09\linewidth]{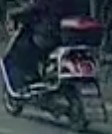} 
                \includegraphics[height=.09\linewidth]{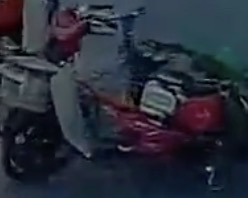}\\ \hline
            \textit{open-tricycle} & 8 &
                \includegraphics[height=.09\linewidth]{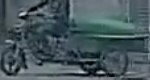} 
                \includegraphics[height=.09\linewidth]{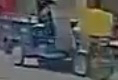} 
                \includegraphics[height=.09\linewidth]{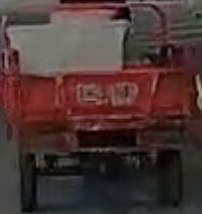}\\ \hline
            \textit{closed-tricycle} & 9 &
                \includegraphics[height=.09\linewidth]{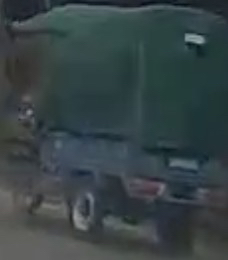} 
                \includegraphics[height=.09\linewidth]{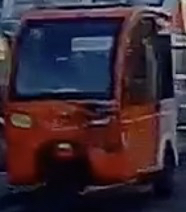} 
                \includegraphics[height=.09\linewidth]{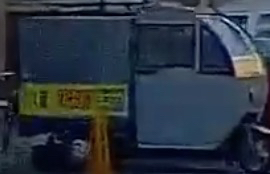}\\ \hline
            \textit{forklift} & 10 &
                \includegraphics[height=.09\linewidth]{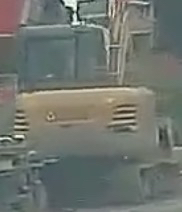} 
                \includegraphics[height=.09\linewidth]{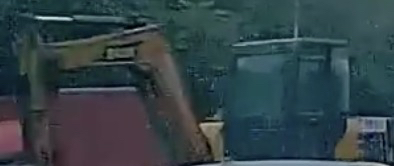} 
                \includegraphics[height=.09\linewidth]{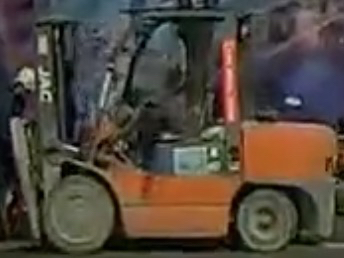}\\ \hline
            \textit{large-block} & 11 &
                \includegraphics[height=.09\linewidth]{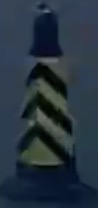} 
                \includegraphics[height=.09\linewidth]{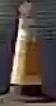} 
                \includegraphics[height=.09\linewidth]{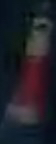}\\ \hline
            \textit{small-block} & 12 &
                \includegraphics[height=.09\linewidth]{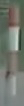} 
                \includegraphics[height=.09\linewidth]{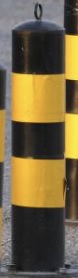} 
                \includegraphics[height=.09\linewidth]{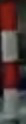}\\
            \bottomrule
        \end{tabular}
    \end{center}
    \caption{Examples of the 12 annotated classes of objects.}
    \label{tab:class-example}
\end{table*}

\end{document}